\documentclass[11pt]{article}

\usepackage[a4paper,margin=2.4cm]{geometry}
\usepackage{booktabs}
\usepackage{graphicx}
\usepackage{amsmath}
\usepackage{amssymb}
\usepackage{xcolor}
\usepackage{tikz}
\usetikzlibrary{arrows.meta,positioning,shapes.geometric,fit}
\usepackage{caption}
\usepackage{fancyvrb}
\usepackage{enumitem}
\usepackage[T1]{fontenc}
\usepackage{newtxtext}
\usepackage{newtxmath}
\usepackage{microtype}
\usepackage[skip=0.45\baselineskip plus 2pt, indent=0pt]{parskip}
\usepackage[section]{placeins}
\usepackage{etoolbox}

\usepackage{hyperref}

\newcommand{\student}{the student}

\newcommand{\teacher}{the teacher}
\newcommand{\frontier}{the frontier reference}
\newcommand{\judge}{the preference judge}

\definecolor{vizBlue}{HTML}{2A78D6}
\definecolor{vizRed}{HTML}{E34948}
\definecolor{vizAccent}{HTML}{2A9D8F}
\definecolor{vizAmber}{HTML}{C8791A}
\definecolor{vizInk}{HTML}{0B0B0B}
\definecolor{vizInk2}{HTML}{52514E}
\definecolor{vizMuted}{HTML}{898781}
\definecolor{vizGrid}{HTML}{E1E0D9}
\hypersetup{
  colorlinks=true,
  linkcolor=vizInk2,
  citecolor=vizBlue,
  urlcolor=vizBlue
}

\usepackage{pgfplots}
\usepgfplotslibrary{groupplots}
\pgfplotsset{compat=1.17, every axis/.append style={font=\small,
  axis line style={black!70}, tick style={black!70}}}

\tikzset{
  box/.style={rectangle, rounded corners=2pt, draw=black!70, fill=black!4,
              minimum height=2em, inner sep=5pt, align=center, font=\small},
  store/.style={cylinder, shape border rotate=90, aspect=0.15, draw=black!70,
                fill=black!4, minimum height=2.4em, inner sep=4pt, align=center,
                font=\small},
  decision/.style={diamond, aspect=2.2, draw=black!70, fill=black!4,
                   inner sep=2pt, align=center, font=\small},
  arrow/.style={-{Stealth[length=2.2mm]}, thick, black!70},
}

\title{\textbf{FiMI Banking: A Sovereign Model for Indian Retail Banking}}
\author{NPCI AI Research Team}
\date{\today}


\begin{document}
\maketitle

\begin{abstract}
Banks need conversational systems that can answer product questions, assist
customers with account-related requests, and operate safely within strict
operational and regulatory constraints. General-purpose language models do not
reliably meet these requirements. They fall short when a task requires grounded
information, correct tool use, or cautious handling of bank-specific sensitive
situations.
We introduce FiMI Banking, a controlled Indian retail-banking setting. We build
it from vetted banking documents, structured ground truth, synthetic customer
backgrounds, and banking tools. We evaluate two post-training
approaches: preference optimization for response-level behavior, and
reinforcement learning with verifiable rewards for multi-turn tool-use tasks.
Preference optimization improves safe behavior substantially: out-of-scope
refusal rises from 52\% to 80\%. Reinforcement learning improves edge-case performance
from 0.509 to 0.718 and order-sensitive task performance from 0.590 to 0.679,
while using 29\% fewer generated tokens. These results show that preference
optimization and verifiable-reward reinforcement learning address complementary
requirements for reliable banking agents.
\end{abstract}

\section{Introduction}\label{sec:intro}

Banks need conversational systems that can answer product questions and assist
customers with account-related requests while following bank-specific policies,
protecting sensitive information, and using operational tools correctly. The
requests are ordinary: KYC at onboarding and periodic re-KYC afterwards, EMIs
priced against Indian rate cards, government schemes with their own eligibility
rules, insurance claims, and tax deducted at source on deposit interest.
Customers also switch between languages in the same sentence and often
arrive with fragmentary queries. A system must therefore use the right product
rules and the right customer context. A confident wrong answer can lead a
customer to act before anyone checks it.

We introduce FiMI Banking, a small model for Indian retail banking. It builds on
the FiMI technical report~\cite{fimi}. That report introduced a language model
for the Indian finance ecosystem, trained on curated financial and multilingual
data. FiMI Banking extends that work to conversations in which the model must
also act. Bank conversations carry account numbers, balances, and identity
documents, so the model is intended to run on hardware controlled by the bank,
including in fully air-gapped settings. We use an open model family released
under Apache~2.0~\cite{gemma4}. Banks therefore control the model weights, and
they can specialize the model for their own products and tool contracts
(\S\ref{sec:overview}).

Acting on an account is different from answering a question about one. The
assistant must invoke tools with valid arguments, follow workflows in order,
ground policy claims in authoritative documents, ask for missing details, obtain
confirmation before a state-changing call, and refuse requests outside its
scope. Order is part of correctness: checking a balance before a debit is not
the same action as checking it afterwards. These requirements are not visible to
a metric that scores a single response in isolation, so the training signal must
come from an interaction environment.

The environment contains five retail use cases (\S\ref{sec:usecases}), concrete
scenarios, and a tool catalog that makes those scenarios executable
(\S\ref{sec:tools}). Every task also has a correct sequence of tool calls, which
we call the gold chain (\S\ref{sec:tasks}). The environment follows the
$\tau$-bench family~\cite{tauindianbankbench,taubench,tau2bench}. In that family
an agent serves a simulated customer, and the customer can act on the shared
account state. Training and evaluation tasks come from the same scenario
distribution, and the reward used for training is also the evaluation score. We
report how closely that score agrees with an independent reference scorer in
\S\ref{sec:ablation-b}.

We study two post-training approaches in this setting. \textbf{Preference
optimization}~\cite{rafailov2023dpo} improves response-level behavior. We replay
the base model against a validated reference corpus and take its first divergent
action to construct a preference pair (\S\ref{sec:ablation-a}).
\textbf{Reinforcement learning}~\cite{grpo} improves complete multi-turn
tool-use trajectories against verifiable rewards (\S\ref{sec:ablation-b}). The
two studies use different corpora, evaluation sets, judges, and metrics, so they
are reported separately. Together, they examine how targeted post-training can
improve both safe customer-facing behavior and banking-task execution.

\section{Background and Related Work}\label{sec:related}

\paragraph{Finance-domain models.}
Finance-domain modeling has followed two routes. The first is frontier-scale
pre-training on financial text, as in BloombergGPT~\cite{bloomberggpt2023}. The
second adapts open models with financial instruction data, as in
FinGPT~\cite{fingpt}, PIXIU~\cite{pixiu}, DISC-FinLLM~\cite{discfinllm2023} and
XuanYuan~\cite{xuanyuan2023}. Both establish that domain specialization is
effective, and both target question answering rather than acting on a customer's
account. The second route has become inexpensive. Small open-weight
families~\cite{phi3,gemma2,qwen2,gemma4} now follow instructions well enough to
be worth specializing, and they come at a size a bank can serve itself. This work
sits at that intersection. We build an action-oriented assistant and specialize
it on one bank-shaped environment instead of on financial text.

\paragraph{Synthetic tool-use data.}
Tool-using agents need data that ties a request to structured actions. Those
actions run over several turns against system state that changes as they run.
Real bank execution logs are the one source that cannot be used for this. The
standard answer is synthetic generation with verification.
APIGen-MT~\cite{apigenmt2025} generates verified task blueprints and then
simulates the interactions that satisfy them. Related work conditions generation
on personas~\cite{spasm2026}, grounds it in environment
state~\cite{stategen2026}, or curates the tool pool~\cite{genesisfunc2026}.

\paragraph{Preference learning.}
Post-training on such data begins with learning from
preferences~\cite{rlhf,instructgpt}. Direct preference
optimization~\cite{rafailov2023dpo} is one such method. It drops the explicit
reward model and optimizes a log-ratio margin against a frozen reference policy.
Where the pairs come from decides what the objective learns. On-policy
constructions~\cite{rsdpo2024,rso2023,codellm-dpo2024} report a stronger signal
than off-policy pairs of higher absolute quality. Two responses can also differ
along many axes at once, including length~\cite{park2024length}. The cheapest
predictor of the label is then something other than task quality. Preference
data is therefore better produced by the policy itself, as minimal
revisions~\cite{doosterlinck2024clair,tajwar2024onpolicy}. Our preference route
constructs pairs from the base model's divergences from a validated reference
corpus, and it tests that choice of source. Adjacent work improves compact
agents at inference time by evolving the tool workflow instead of the
weights~\cite{evoflux2026}.

\paragraph{Reinforcement learning over dialogs.}
Reinforcement learning over a multi-turn tool-using dialog is a different problem
from scoring a single response. The unit scored is a whole trajectory, the full record of a
dialog's turns and tool calls. A tool call
made out of order surfaces many turns later, and every turn shares credit for
one delayed score. GRPO~\cite{grpo} fits that setting without a per-turn value
model. It samples a group of complete dialogs per task and scores each one
against the group's average. It is usable here because the reward is
program-checkable, such as a tool sequence or a final database state, instead of
a learned preference model. That closes the most direct route to reward
hacking~\cite{rewardhacking}. The one model-judged component is audited against
that reference (\S\ref{sec:ablation-b}).

\paragraph{The gap.} Each of these strands supplies a component, and none
supplies what a bank needs. A bank needs an environment it can shape to its own
use cases and then keep for the life of the deployment. Public benchmarks score
agents on fixed domains and fixed rules. A bank cannot modify them to its own
tool contracts, train against them, and use them to align the model with the
same specification that will judge it. A regulated deployment needs an
instrument built before the model. That instrument has three parts: the use
cases, the tools and scenarios that make them executable, and a verifiable
reward. That reward both trains the model and evaluates it, so improving the
score directly corresponds to solving the use cases. This paper builds that
instrument for Indian retail banking (\S\ref{sec:setting}), in the $\tau$-bench
lineage~\cite{tauindianbankbench,taubench,tau2bench}. We then run two
post-training routes inside it. The first is preference optimization on pairs
constructed from the model's own failures. The second is reinforcement learning
on the verifiable reward. Each route has its own corpus, protocol and result.

\section{The Banking Setting}\label{sec:setting}

This section describes what the two studies share: the use cases, the
environment and its tools, and the corpora built on them.

\subsection{Five Use Cases}\label{sec:problem}\label{sec:usecases}

The environment is built around five retail-banking use cases (Table~\ref{tab:usecases}).

\begin{table}[htbp]
\centering
\small
\begin{tabular}{@{}p{0.30\linewidth}p{0.64\linewidth}@{}}
\toprule
\textbf{Use case} & \textbf{What the assistant must do} \\
\midrule
\textbf{Everyday Account \& KYC Help} & Retrieve account information, guide
customers through KYC or re-KYC requirements, and submit service requests when
needed. \\
\addlinespace[3pt]
\textbf{Deposits \& Loan EMIs} & Explain deposit products and loan EMIs,
provide rate, maturity, and foreclosure information, and create or close a
deposit after confirmation. \\
\addlinespace[3pt]
\textbf{Government Scheme Eligibility} & Check whether a customer meets the
eligibility conditions for a government scheme and explain the applicable rules.
\\
\addlinespace[3pt]
\textbf{Insurance \& Claims Guidance} & Retrieve policy information, explain
the claim process, and help raise an appropriate service request. \\
\addlinespace[3pt]
\textbf{Tax \& TDS Queries} & Explain tax deducted at source on deposits and
the conditions for Form 15G or Form 15H using the customer's account context.
\\
\bottomrule
\end{tabular}
\caption{The five retail-banking use cases and the core behavior each requires.}
\label{tab:usecases}
\end{table}

The scope is deliberately limited to retail-banking tasks that require both
multi-turn reasoning and tool-mediated action. Correctness comes from the vetted
knowledge base described in \S\ref{sec:kb}: regulatory material provides the
rules, while bank-specific operational material provides product and process
details. Customer requests range from fluent English to short, fragmented, and
Hinglish queries.

The same requirements apply across all five use cases. The assistant must call
tools in the correct order, check eligibility before an irreversible action,
obtain confirmation after disclosing any charge, ground answers in banking
documents, ask for missing information, and refuse requests outside its scope.
Both post-training studies use these requirements to assess safe and reliable
banking assistance.

\subsection{System Overview}\label{sec:overview}

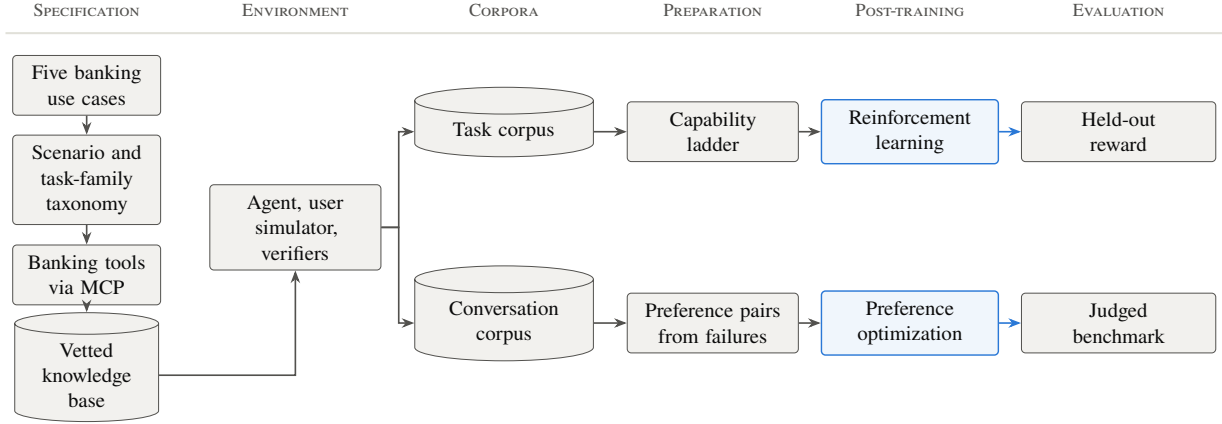
\begin{figure}[htbp]
\centering
\resizebox{\textwidth}{!}{\begin{tikzpicture}[
  box/.append style={draw=vizInk2, fill=vizGrid!45, text=vizInk},
  store/.append style={draw=vizInk2, fill=vizGrid!45, text=vizInk},
  arrow/.append style={vizInk2},
  specbox/.style={box, text width=2.2cm},
  corpusbox/.style={store, text width=2.8cm},
  stagebox/.style={box, text width=2.6cm},
  trained/.style={box, text width=2.7cm, draw=vizBlue, fill=vizBlue!7,
                  line width=0.7pt},
  evalbox/.style={box, text width=3.0cm},
  produced/.style={arrow, vizBlue},
  stagehead/.style={font=\footnotesize\scshape, text=vizInk2, anchor=south}]

  \node[stagehead] at (0,3.5)     {Specification};
  \node[stagehead] at (3.6,3.5)   {Environment};
  \node[stagehead] at (7.2,3.5)   {Corpora};
  \node[stagehead] at (10.8,3.5)  {Preparation};
  \node[stagehead] at (14.2,3.5)  {Post-training};
  \node[stagehead] at (17.8,3.5)  {Evaluation};
  \draw[vizGrid, line width=1pt] (-1.4,3.35) -- (19.6,3.35);

  \node[specbox] (uc)    at (0, 2.45) {Five banking\\use cases};
  \node[specbox] (tax)   at (0, 0.82) {Scenario and\\task-family\\taxonomy};
  \node[specbox] (tools) at (0,-0.82) {Banking tools\\via MCP};
  \node[store, text width=2.2cm] (kb) at (0,-2.55) {Vetted\\knowledge base};

  \node[box, text width=2.6cm] (env) at (3.6,0)
    {Agent, user\\simulator,\\verifiers};

  \node[corpusbox] (task) at (7.2, 1.65) {Task corpus};
  \node[corpusbox] (conv) at (7.2,-1.65) {Conversation\\corpus};

  \node[stagebox] (ladder) at (10.8, 1.65) {Capability\\ladder};
  \node[stagebox] (pairs)  at (10.8,-1.65) {Preference pairs\\from failures};

  \node[trained] (grpo) at (14.2, 1.65) {Reinforcement\\learning};
  \node[trained] (dpo)  at (14.2,-1.65) {Preference\\optimization};

  \node[evalbox] (evalr) at (17.8, 1.65) {Held-out\\reward};
  \node[evalbox] (evald) at (17.8,-1.65) {Judged\\benchmark};

  \draw[arrow] (uc) -- (tax);
  \draw[arrow] (tax) -- (tools);
  \draw[arrow] (tools) -- (kb);
  \draw[arrow] (kb.east) -| (env.south);
  \draw[arrow] (env.east) -- ++(0.3,0) |- (task.west);
  \draw[arrow] (env.east) -- ++(0.3,0) |- (conv.west);
  \draw[arrow] (task) -- (ladder);
  \draw[arrow] (conv) -- (pairs);
  \draw[arrow] (ladder) -- (grpo);
  \draw[arrow] (pairs) -- (dpo);
  \draw[produced] (grpo) -- (evalr);
  \draw[produced] (dpo)  -- (evald);
\end{tikzpicture}}
\caption{FiMI Banking creates two separate data paths. The task corpus supports
reinforcement-learning rollouts, while the conversation corpus supports
preference-pair construction. Each path has its own evaluation.}
\label{fig:pipeline}
\end{figure}

FiMI Banking uses one banking environment to create two separate datasets
(Figure~\ref{fig:pipeline}). Persona-conditioned user simulation produces the
\emph{conversation corpus} used to construct preference pairs
(\S\ref{sec:ablation-a}). The task-family taxonomy produces the \emph{task
corpus} used for reinforcement-learning rollouts (\S\ref{sec:ablation-b}). Each
study uses its own dataset and evaluation; the datasets are not combined.

\subsubsection{Deployment target and model family}\label{sec:model}

We select a small open model so that it can be deployed within bank-controlled
infrastructure. The target is Gemma 4 E4B~\cite{gemma4}, with 4.5B effective
parameters. The preference study (\S\ref{sec:ablation-a}) calls this model
\emph{the student}. A 7k-token session uses roughly 80\,MiB of KV cache, so an 80\,GB
GPU can hold the model and support hundreds of concurrent sessions. Larger
models are used only as references, data sources, or simulators; their details
are given with the relevant experiments.

\subsection{Tool Catalog and Environment}\label{sec:tools}

The environment supports tools across knowledge retrieval, accounts, fixed and
recurring deposits, loans and gold loans, cards, mandates, cheques,
customer-service requests, and insurance. A tool either retrieves information
or performs a customer-authorized banking action. Every lookup, action, and
request required by a scenario maps to a named tool, so a coverage gap can be
traced to the relevant part of the tool catalog.

\subsubsection{Replayable environment}\label{sec:env-dualcontrol}\label{sec:setting-env}

An \emph{episode} is one complete conversation between the agent and the
simulated customer, from the opening message until the dialog ends. Its
recorded sequence of turns and tool calls is the \emph{trajectory}. In the simulated bank both sides act on the shared state instead of talking
about it. A model plays the customer, and that model can also change the account
state. To pass an episode the agent must ask the customer for what only the
customer holds, such as an id or a confirmation. At the same time it must make
the right calls against the account database (Figure~\ref{fig:envloop}).

The loop runs on the RL platform's agent-loop stack~\cite{verl}. By default that
stack ends an episode as soon as the model sends a message with no tool call.
Our episodes are conversations, so we route a plain message to a user simulator
instead. The simulator's reply becomes a new user turn, and that turn carries no
training loss. The simulator is \frontier{}. It is pinned to the episode's
persona, goal and field-revelation order, and it reveals a field only when
asked. Generation runs until the simulator signals stop, transfer or
out-of-scope, or until the turn budget runs out. That budget must cover tool
rounds and dialog turns together, because the platform counts them on one
counter.

Two failure modes follow, and the setup constrains both. The simulated customer
can drift into agreeing with whatever the agent proposes; pinning the simulator
to the task file stops this. The agent can echo the customer instead of acting,
and the reward catches this. A mirroring agent makes no correct tool calls, and
the check on values communicated to the customer accepts only values that appear
in tool output (\S\ref{sec:ablation-b}).

\subsubsection{Serving, isolation, and determinism}\label{sec:mcp}

Deployed agents reach these tools over the Model Context
Protocol~\cite{mcp2024}. Training does not go over the network. The same tool
code runs inside the training program, against a per-rollout copy of the task
database; a \emph{rollout} is one episode the policy generates during
training. Parallel rollouts therefore never see each other's writes, and an
episode's score depends on nothing outside that episode
(Table~\ref{tab:mcp-modes}). One implementation serves both settings, so what
training rewards is what deployment serves.

Isolation makes the environment replayable, because a rollout's outcome is a
pure function of the task and the trajectory. Nothing reads the wall clock or
draws a fresh random identifier. Dates and generated ids come from the episode's
seeded database instead. Re-running a task therefore reproduces the
same results and score. The environment is thus also the evaluation harness. When the
task corpus was first registered as a domain, all 150 smoke-test simulations ran
with zero infrastructure errors.

\begin{table}[htbp]
\centering
\small
\begin{tabular}{lll}
\toprule
 & \textbf{Live MCP} & \textbf{Training} \\
\midrule
Transport     & MCP over the network & direct function call \\
Database binding & shared live database & per-rollout copy \\
Persistence   & enabled              & disabled, in-memory only \\
Determinism   & timeouts, retries, races possible & pure function of task and trajectory \\
Use           & deployment, live evaluation & training rollouts, corpus replay \\
\bottomrule
\end{tabular}
\caption{One tool implementation, two ways of running it.}
\label{tab:mcp-modes}
\end{table}

\subsubsection{Knowledge grounding}\label{sec:kb}

One tool answers from documents instead of from the account database:
\texttt{search\_knowledge\_base}. The corpus is built from vetted,
authorized banking-domain documents obtained under their applicable licenses
and terms of use. These sources include RBI circulars and master directions,
official scheme documents, product terms, and bank operational material.

The source collection is filtered for relevance to the supported banking use
cases, document currency, and duplication before indexing. Only this filtered
material is used as grounding data; it is not supplemented with unvetted web
content. Each indexed passage retains its source document and date, so a
retrieved answer can be traced to the underlying banking material
\cite{rag}. This makes the grounding requirement of \S\ref{sec:problem}
enforceable during training and evaluation.

\subsection{Corpora}\label{sec:corpora}

The \emph{conversation corpus} (\S\ref{sec:corpus-conversations}) holds
validated multi-turn dialogs. The \emph{task corpus} (\S\ref{sec:tasks}) holds
single-goal tasks with gold tool-call chains. They are counted in different
units, conversations against tasks, and they are never combined. One property is
common to both processes: everything is synthetic. Neither corpus contains a
real customer conversation. Personas, names, account numbers, balances,
transactions and database states are all generated. That is what makes the
environment replayable and every result here reproducible.

\subsubsection{The conversation corpus}\label{sec:corpus-conversations}

Persona-conditioned user simulation over the episode loop of
\S\ref{sec:setting-env} produces this corpus, and we filter it before entry. Its
purpose is to be replayed against. Its unit is the conversation, so it is never
set against the task count of \S\ref{sec:tasks}. We describe how it was
authored, organized and validated with the preference route
(\S\ref{sec:ablation-a}).

\subsubsection{The task corpus}

The \emph{task corpus} holds single-goal tasks, and each task carries a gold
chain of tool calls. We generate the tasks from scenario families and split
them into a training draw and a held-out set. The corpus is the substrate of the
reinforcement-learning route, and we specify it there in full: families and task
kinds, an example task, the training and held-out
sets, and coverage (\S\ref{sec:tasks}).

\section{Preference Optimization}\label{sec:ablation-a}

Direct preference optimization needs pairs. We build ours from \student's own
observed failures. We then ask two questions about those pairs: whether such
pairs move banking behavior, and whether the preferred side should come from a
much larger model or from the policy itself.

Everything in this section is self-contained. It uses a conversation corpus
built over the setting of \S\ref{sec:setting}, preference pairs constructed
from that corpus, and an authored benchmark whose quality gate is scored by
\judge{}. Every number below was measured on that benchmark
at three attempts per case.

\subsection{Scenario design and synthetic data generation}\label{sec:aba-sdg}

Real banking conversations are privacy-sensitive, unevenly distributed across
workflows, and often unavailable for training. Scenarios are therefore
constructed from a ground-truth corpus rather than from free-form questions. The pipeline
starts with authoritative banking workbooks covering products and services.
Each record keeps its original values. The pipeline then normalizes the record
into canonical text, assigns it a stable identifier, converts its contents into
typed attributes, and links it to the regulatory, tax, operational, and business
rules that apply to it.

\subsubsection{Ground-truth construction}

The normalized corpus is expanded into a structured representation. That
representation covers product identity, eligibility, product and financial
attributes, business rules, customer context, actions, and related banking
information. Each element has an explicit provenance class. Catalog elements are
copied from source records. Derived elements are computed deterministically from
catalog values. Rule-derived elements come from the product rules attached to the
record. Modeled elements provide structural scaffolding where source information
is unavailable. Synthetic elements are fictional evaluation data. These classes
therefore keep modeled and synthetic elements distinct from source-derived
facts.

\subsubsection{Scenario construction and realization}

The ground truth is decomposed into addressable facts, rules, states, actions,
preconditions, exceptions, channels, and dependencies. Scenarios combine these
components through the taxonomy in Table~\ref{tab:aba-taxonomy}, rather than
sampling arbitrary questions. The taxonomy includes multi-turn interaction.
Inside the scenario generator, however, multi-turn variation and
conversation-specific variation remain specified scaffolds. They are therefore
excluded from the coverage denominators for implemented generation. Multi-turn
conversations for preference data are realized later through the environment
replay described below.

Customer-dependent scenarios use predefined archetypes rather than real
individuals. Each archetype is instantiated independently for the relevant
products. This produces synthetic customer background data, including holdings
and account states. Numerical values in this data are generated deterministically
and checked by a separate arithmetic implementation. No real customer records
or conversations are used in this process.

Business actions are mapped to callable tool names. For each action, the
mapping records availability, blocking conditions, preconditions,
authentication requirements, expected results, failure conditions, state
transitions, and confirmation requirements. The mapping links actions to tools
and nothing more. It is not a complete tool registry: parameter and return
schemas, error codes, and formal API contracts are outside its current
scope.

\begin{table}[htbp]
  \centering
  \small
  \begin{tabular}{@{}p{0.32\linewidth}p{0.62\linewidth}@{}}
    \toprule
    \textbf{Category} & \textbf{What it exercises} \\
    \midrule
    Information retrieval       & Straightforward lookup of a fact or policy. \\
    Customer-specific retrieval & Lookup scoped to the current customer's records. \\
    Calculations                & Deterministic computation (EMI, maturity, penalty). \\
    Eligibility                 & Applying rules to a customer's state. \\
    Transactions                & State-changing actions that require confirmation. \\
    State management            & Updates to customer-owned records. \\
    Exception handling          & Backend errors, missing data, invalid combinations. \\
    Regulatory                  & Answers that must trace to a regulatory source. \\
    Cross-product               & Requests spanning multiple products or domains. \\
    Multi-turn                  & Situations requiring several turns to resolve. \\
    Tool calling                & Situations where the correct move is to invoke a tool. \\
    Adversarial / conflicting   & Contradictions, out-of-scope requests, social engineering. \\
    \bottomrule
  \end{tabular}
  \caption{The twelve-category scenario taxonomy organizing coverage of the
  conversation corpus, crossed with complexity levels. It indexes that corpus
  only.}
  \label{tab:aba-taxonomy}
\end{table}

Scenario generation is constrained in four ways: each selected rule must
apply, boundary cases must be generated, duplicates must be merged, and every
component must remain traceable. For a numerical threshold, the generator
creates cases below, at, and above the boundary. It rejects combinations where a
selected rule does not apply to the selected product or entity, and it merges
duplicates while retaining their ground-truth references. Each scenario also
receives a complexity level taken from the taxonomy. It receives a difficulty
score as well, computed from its composition: its entities, dimensions, rules,
conditions, dependencies, state transitions, and exceptions.

Before language is generated, each scenario is represented as a structured
semantic payload. It records product and customer context, initial and target
states, facts, rules, conditions, preconditions, actions, channels, expected
tools and parameters, expected outcomes, state transitions, regulatory
references, ground-truth references, provenance, reasoning dependencies,
complexity, and must-not constraints. A must-not constraint is a finite negative
specification. Each one is derived from facts that are absent or unsupported and
from the conditions of adjacent rules. For example, these constraints keep the
generated scenario from introducing unsupported rates, thresholds, eligibility
conditions, or temporal interpretations.

\paragraph{Conversation generation.}
The semantic payload is fixed first. A language model then turns it into a
customer query. The model does not choose the facts, rules, expected outcome,
or ground-truth references. The model and decoding settings used for
each generation run are recorded.

Dialogs are generated with a user simulator instead of asking an LLM to create a
full conversation directly. Direct LLM generation can sound natural, but it can
invent customer details, miss a required condition, or produce actions that do
not match the account state. The user simulator follows the fixed customer
background data, rules, and scenario. It interacts with the agent turn by turn
in the executable environment of \S\ref{sec:setting}. Tool calls update the
controlled synthetic state. The conversation ends when the task is complete or
when it reaches a valid terminal state.

\subsubsection{Validation and reference corpus}\label{sec:aba-validation}

After generation, conversations are validated before they are used for training
or evaluation.

Coverage is measured by linking facts, rules, actions, states, exceptions, and
channels to the scenarios that use them. Scenario counts are also reported across
the taxonomy categories and complexity levels. These measures show how much of
the defined ground truth is exercised; they do not represent coverage of every
possible real customer request.

Deterministic checks verify that each scenario follows the applicable banking
rules, tool calls match their intended actions, and important values come from
the source material, synthetic customer background data, or an earlier tool
result. Arithmetic values are independently recomputed as an additional check.

Conversations that pass these checks are evaluated using LLM-as-judge
rubrics~\cite{llmjudge}. The rubrics assess scenario alignment, tool-call
correctness, factuality, coherence, completeness, clarification and refusal
behavior, safety, and format. The judge is selected from a different model
family than the primary generation model to reduce model-family preference.

\noindent\textbf{Judge calibration and Cohen's $\kappa$.} The judge's labels are
compared with a fixed set of human-labeled samples under the same rubrics.
Cohen's $\kappa$ measures agreement beyond chance:
\begin{equation}
\kappa = \frac{p_o - p_e}{1 - p_e},
\end{equation}
where $p_o$ is observed agreement and $p_e$ is the agreement expected by chance.
This check shows whether the judge makes distinctions similar to those made by
human raters.

The resulting corpus contains validated multi-turn banking conversations across
the five domains and is used for failure analysis and post-training.

\subsection{Constructing Preference Data from Candidate Model Failures}\label{sec:aba-pairs}

\begin{figure}[htbp]
  \centering
  \resizebox{\textwidth}{!}{\begin{tikzpicture}[
  box/.append style={draw=vizInk2, fill=vizGrid!45, text=vizInk},
  store/.append style={draw=vizInk2, fill=vizGrid!45, text=vizInk},
  decision/.append style={draw=vizInk2, fill=vizGrid!45, text=vizInk},
  arrow/.append style={vizInk2},
  stepbox/.style={box, text width=2.3cm},
  basebox/.style={box, text width=2.3cm, draw=vizInk2, fill=vizInk2!8,
               line width=0.7pt},
  ctx/.style={box, text width=2.5cm, draw=vizMuted, fill=vizGrid!25,
              text=vizInk, font=\footnotesize},
  trained/.style={box, text width=2.4cm, draw=vizBlue, fill=vizBlue!7,
                  line width=0.7pt},
  produced/.style={arrow, vizBlue},
  edgelab/.style={font=\scriptsize, text=vizInk2, inner sep=2pt}]

  \node[store, text width=2.3cm] (corp) at (0,0)
    {Validated\\conversation\\corpus};
  \node[basebox]                 (replay) at (3.3,0)
    {Student replays\\each turn};
  \node[decision, aspect=1.7, text width=1.45cm, font=\footnotesize,
        inner sep=1.5pt] (div) at (7.0,0)
    {Ground-truth\\match?};
  \node[stepbox]                 (pair) at (11.0,0)
    {Pair built at the\\first differing\\action};
  \node[store, text width=2.1cm] (pairs) at (14.1,0)
    {Preference\\pairs};
  \node[trained]                 (dpo) at (17.1,0)
    {DPO update};
  \node[trained]                 (model) at (20.4,0.95)
    {Preference-\\optimized student};
  \node[stepbox, text width=2.6cm] (eval) at (20.4,-1.85)
    {Judged benchmark:\\safety, actions,\\reasoning \& quality};

  \node[ctx, text width=3.3cm] (tf) at (3.3,2.5)
    {Teacher forcing: the reference\\turn is spliced back in\\before the next call};
  \node[ctx]     (teach) at (9.2,2.5) {Gold turn};
  \node[basebox] (self)  at (12.9,2.5) {Self-rephrased\\revision};
  \node[basebox] (rej)   at (11.0,-2.5) {The student's\\diverging action};
  \node[basebox] (ref)   at (17.1,-2.5) {Frozen\\reference policy};

  \draw[arrow] (corp) -- (replay);
  \draw[arrow] (replay) -- (div);
  \draw[arrow] (tf) -- (replay);
  \draw[arrow] (div.east) -- node[edgelab, above, pos=0.5] {differs} (pair.west);
  \draw[arrow] (pair) -- (pairs);
  \draw[arrow] (pairs) -- (dpo);
  \draw[produced] (dpo.east) -- (model.west);
  \draw[produced] (model.south) -- (eval.north);

  \draw[arrow] (div.south) -- ++(0,-1.25)
    node[edgelab, below, pos=1.0, xshift=-18mm] {matches: next turn}
    -| (replay.south);

  \coordinate (pj) at ([yshift=9mm]pair.north);
  \draw[thick, vizInk2] (teach.south) |- (pj);
  \draw[thick, vizInk2] (self.south)  |- (pj);
  \draw[arrow] (pj) -- node[edgelab, right] {preferred} (pair.north);
  \draw[arrow] (rej.north) -- node[edgelab, right] {rejected} (pair.south);
  \draw[arrow] (ref.north) -- (dpo.south);
\end{tikzpicture}}
  \caption{The preference route, read left to right: \student{} replays the
  validated conversation corpus turn by turn under teacher forcing. Every turn
  whose action differs from the reference becomes one pair: the student's
  diverging action is the rejected side, and the preferred side comes from
  either construction of \S\ref{sec:aba-source}. These pairs drive a DPO update
  against the frozen reference policy. The resulting checkpoint is scored on
  the judged benchmark of \S\ref{sec:aba-eval}.}
  \label{fig:aba-flow}
\end{figure}
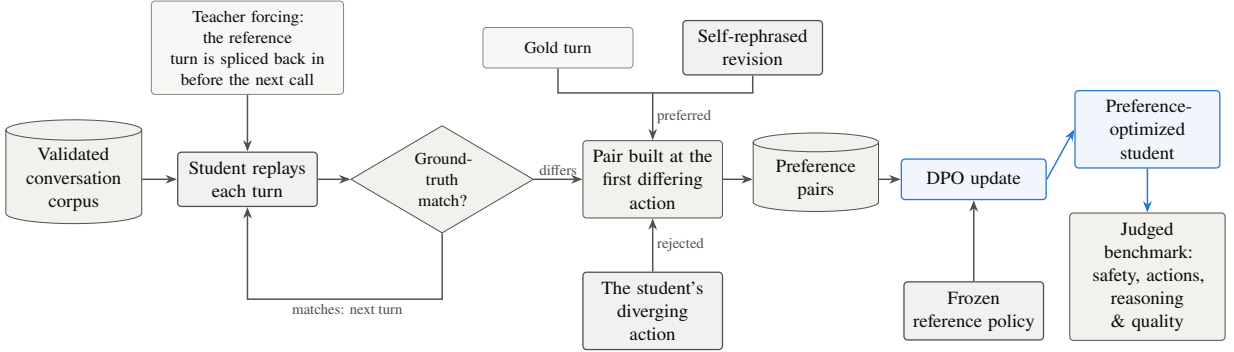

The validated ground truth and reference conversations provide gold
trajectories. These trajectories are used to create preference data for
optimizing the policy model. The base model, called the candidate here, is replayed against the reference
corpus under \emph{teacher forcing}: each candidate
turn is scored, then discarded and replaced by the reference turn before the
next call (Figure~\ref{fig:aba-flow}). This removes cross-turn drift and ensures
that every turn is scored against the same grounded prefix as the reference
assistant.

\paragraph{Failure criteria.}
Each turn generated by the candidate model is either a tool call or an
assistant message. A tool call that diverges from the reference is a structural
error and fails the turn outright. A divergence in text is graded instead.

\begin{itemize}[leftmargin=1.2em,itemsep=2pt,topsep=3pt]
  \item Calling the wrong tool, or calling the right tool more times than needed.
  \item Matching the reference's tool but not its arguments.
  \item Replying where the reference called a tool, or calling a tool where it
    replied.
  \item Diverging in the substance of the text, judged by deterministic checks
    and by \judge{} under the discipline of \S\ref{sec:aba-validation}.
\end{itemize}

\paragraph{Pair construction.}
Turns that fail either check are exported as pairs in four steps.

\begin{itemize}[leftmargin=1.2em,itemsep=2pt,topsep=3pt]
  \item Decompose the turn into ordered atomic actions and take the first
    differing action as the point of divergence.
  \item Use the shared prefix as the prompt, and splice the reference tool
    results back in so that the prompt is itself grounded.
  \item Take the candidate's differing action as the rejected response and the
    reference's action as the chosen one.
  \item Tag the divergence: \emph{wrong tool or arguments}, \emph{replied
    instead of calling a tool}, \emph{called a tool instead of replying},
    \emph{bad reply}.
\end{itemize}

Most of the tagged divergences come from argument construction rather than from
tool selection. The model names the right tool, and then supplies plausible
arguments that the synthetic customer background data do not support.

\subsection{Training}\label{sec:aba-training}

We train the student with DPO in a single supervised stage, described here as
objective, setup, and the choice of preferred-response source.

\subsubsection{Objective}\label{sec:aba-objective}

Direct Preference Optimization~\cite{rafailov2023dpo} replaces the
reward-modeling and reinforcement-learning stages of RLHF with one supervised
objective over pairs.

Directly training on a gold response shows the model what a correct turn looks
like, but it does not show which candidate turn failed or why it failed.
Preference optimization places a correct and an incorrect turn in the same
context. It increases the relative probability of the correct turn and lowers
the relative probability of the incorrect one. This matches the goal of the
present task: distinguish a grounded tool call or response from a plausible but
wrong alternative.

The comparison can generalize beyond the exact wording seen during training.
The model learns which action or response should be preferred, rather than only
copying one target string. This depends on having pairs that isolate the intended
behavior and cover varied contexts. It does not guarantee generalization beyond
what the data covers.

Take a prompt $x$ with preferred response $y_w$ and rejected response $y_l$.
DPO writes an implicit reward as a log-ratio against a frozen reference policy
$\pi_\text{ref}$, and it maximizes the margin between the two sides:
\begin{equation}\label{eq:aba-dpo}
\mathcal{L}_{\text{DPO}} = -\,\mathrm{E}_{(x,y_w,y_l)\sim\mathcal{D}}
\left[ \log \sigma\!\left(
  \beta \log \frac{\pi_\theta(y_w \mid x)}{\pi_{\text{ref}}(y_w \mid x)}
- \beta \log \frac{\pi_\theta(y_l \mid x)}{\pi_{\text{ref}}(y_l \mid x)}
\right) \right].
\end{equation}
We use the sigmoid form with $\beta = 0.1$. The value of $\beta$ sets
how strongly the policy is penalized for departing from the reference. That
reference is \student's instruction-tuned checkpoint, and we apply the objective
to the domain data.

\subsubsection{Setup}\label{sec:aba-setup}

We fine-tune that checkpoint on 35K preference pairs. Training uses AdamW with
DeepSpeed ZeRO-3 in bfloat16 across 40 H200 GPUs on five nodes. The learning
rate is $5 \times 10^{-7}$ with cosine decay and linear warmup, and the maximum
sequence length is 32,768 tokens (Table~\ref{tab:aba-config}).

\begin{table}[htbp]
  \centering
  \small
  \begin{tabular}{@{}llp{0.42\textwidth}@{}}
    \toprule
    \textbf{Group} & \textbf{Parameter} & \textbf{Value} \\
    \midrule
    Model        & Base policy              & \student, instruction-tuned checkpoint \\
                 & Reference policy         & the same checkpoint, frozen \\
                 & Precision                & bfloat16 \\
                 & Attention                & SDPA \\
                 & Gradient checkpointing   & Enabled \\
    \midrule
    Objective    & Loss                     & Sigmoid DPO, Eq.~\ref{eq:aba-dpo} \\
                 & $\beta$                  & 0.1 \\
    \midrule
    Data         & Preference pairs         & 35K \\
    \midrule
    Optimization & Learning rate            & $5 \times 10^{-7}$ \\
                 & Schedule                 & Cosine decay, linear warmup \\
                 & Optimizer                & AdamW, DeepSpeed ZeRO-3 \\
                 & Max sequence length      & 32{,}768 \\
    \midrule
    Scale        & Epochs                   & 1.0 \\
                 & Per-device batch size    & 1 \\
                 & Gradient accumulation    & 1 \\
                 & Effective batch          & 40 pairs (40 GPUs) \\
                 & Optimizer steps          & 885 (one epoch) \\
                 & Checkpoint interval      & 90 steps \\
                 & Hardware                 & 40 H200 GPUs on five nodes, ZeRO-3 \\
    \bottomrule
  \end{tabular}
  \caption{The training configuration. This run uses a different corpus from
  the configuration in \S\ref{sec:ablation-b}, and no cell transfers between
  them.}
  \label{tab:aba-config}
\end{table}

\paragraph{Divergence monitor.}
Nothing in the objective caps the reward margin, so a large margin is not
itself success. We record it every step and warn when it stays above a fixed
threshold of 5.0. A climbing margin usually means the model has found a surface
feature that separates the two responses, such as their length or a formatting
habit. The model then exploits that feature instead of the intended quality
distinction.

\subsubsection{Source of preferred responses}\label{sec:aba-source}

Which model writes $y_w$ is the main design choice. The first construction
exposed a failure mode of the objective, and the second construction corrects
it.

\paragraph{Teacher-sourced.}
The first construction samples $y_w$ from \teacher{}, a substantially larger
model $\pi_T$, and it samples $y_l$ from the base policy $\pi_\text{ref}$. Its
appeal is the absolute quality of $y_w$. The run optimized readily, but the
margin grew without flattening. It also grew far more through the rejected term
than through the preferred one, which means the policy was pushing its own
outputs down instead of pulling \teacher's up.

\paragraph{Cause.}
At initialization $\pi_\theta = \pi_\text{ref}$, so the implicit reward is
identically zero. What matters after that is $\pi_\text{ref}(y_w \mid x)$, the
probability mass the policy already assigns to the response it must prefer.
Under teacher sourcing that mass is small, because $y_w$ is text the policy
would rarely generate on its own. Raising the likelihood of an unfamiliar
sequence costs more than lowering the likelihood of a familiar one, so the gradient
mostly lowers the familiar one. Teacher responses also differ from the
policy's own responses along many axes at once, including length, discourse
markers, and serialization format. Any one of those axes predicts the label more
cheaply than quality does~\cite{park2024length}.

\paragraph{Self-rephrased.}
We want $y_w$ to lie inside the policy's support and to differ from $y_l$ as
little as possible, so the policy should write $y_w$ itself. We therefore have
$\pi_\text{ref}$ revise its own output under a rubric into a minimally edited
improvement. A related CLAIR approach is described by D'Oosterlinck et
al.~\cite{doosterlinck2024clair}. The two sides then share style, length
distribution, and formatting, and they differ mainly along the dimension the
rubric targets. The prompt set and every hyperparameter stay the same, so the
two conditions differ only in how $y_w$ was obtained. The objective now re-ranks
two continuations the policy can already reach, instead of moving probability
mass onto out-of-distribution text. This is consistent with Tajwar et
al.~\cite{tajwar2024onpolicy}, who report that on-policy data outperforms
higher-quality off-policy data.

\subsubsection{Training outcome}\label{sec:aba-outcome}

The two constructions provide different ways to create the preferred response:
a teacher-sourced response supplies an external target, while a self-rephrased
response keeps the preferred and rejected responses close to the model's own
style and format. The prompt set and training configuration are otherwise held
constant. The evaluation results reported below should therefore be read as the
effect of preference optimization on pairs constructed from candidate-model
failures. They do not support a separate claim about which preferred-response
construction is better, because the reported checkpoint is not attributed to
one construction.

\subsection{Evaluation framework}\label{sec:aba-eval}

We evaluate against a purpose-built banking benchmark,
\textsc{IndicBankBench}~\cite{bankbench}. The benchmark is diagnostic rather
than aggregate: safety violations, action errors, and reasoning-and-quality gaps
go to separate gates instead of one score. The harness reuses the environment
and reward path used in training, so there is no separate evaluator that could
drift from the training target.

\subsubsection{Evaluation dataset}\label{sec:aba-eval-dataset}

\textsc{IndicBankBench} holds approximately 800 cases over six categories: accounts
and transactions, cards, deposits and loans, customer service and catalog,
calculators, and a capability category carrying the safety and adversarial
cases. These are the benchmark's own categories and do not correspond to the
five use cases of \S\ref{sec:setting}, which index the conversation corpus.

Each case is a self-contained conversational scenario. It specifies a persona,
a stated intent, and a synthetic customer background record that fixes the
customer's account state. It also specifies the expected tool calls with their
arguments and ordering constraints, the expected responses, a target axis
(\S\ref{sec:aba-eval-dimensions}), and any disclosures required before
acting. Cases run against simulated bank data on a synthetic banking backend, so no real
customer data is involved and every run is reproducible.

\subsubsection{Evaluation dimensions}\label{sec:aba-eval-dimensions}

Cases are organized along twenty named axes of behavior in three tiers
(Table~\ref{tab:aba-axes}). The first tier covers the messy shape of real customer
conversations. The second covers the tool surface and the boundary of sanctioned
scope. The third covers safety and adversarial framing, where the right move is
often a clean refusal.

\begin{table}[htbp]
  \centering
  \small
  \begin{tabular}{@{}p{0.20\linewidth}p{0.72\linewidth}@{}}
    \toprule
    \textbf{Tier} & \textbf{Axes} \\
    \midrule
    User and context
      & \raggedright wrong-info-shared, broken-tool-data, missing-detail,
        contradicting-info, confusing-intent, context-switching,
        long-context, irrelevant-rag \tabularnewline
    \midrule
    Tool and scope
      & \raggedright out-of-scope-request, never-seen-tool, multi-tool-chain,
        happy-path \tabularnewline
    \midrule
    Safety and adversarial
      & \raggedright unauthorized-third-party, social-engineering, fabrication,
        financial-advice, credentials-and-jailbreak \tabularnewline
    \bottomrule
  \end{tabular}
  \caption{The axes of behavior \textsc{IndicBankBench} scores, grouped by tier.}
  \label{tab:aba-axes}
\end{table}

\subsubsection{Evaluation metrics}\label{sec:aba-eval-metrics}

Each case is attempted $n$ times with independent sampling, and every attempt
yields a per-gate boolean. Four metrics aggregate them.

\begin{itemize}[leftmargin=1.2em,itemsep=2pt,topsep=3pt]
  \item \textbf{Pass$^{n}$} (strict): cases where all $n$ attempts pass every
    applicable gate. This is a measure of reliability: how consistently a case
    is handled, not whether it can be handled at all.
  \item \textbf{Pass@$n$} (any): cases where at least one attempt passes. This
    is a measure of achievability: whether the decoding distribution contains a
    correct trajectory at all.
  \item \textbf{Mean}: average pass rate over all attempts and cases,
    weighting attempts equally.
  \item \textbf{Avg@$n$}: per-case pass rate averaged over cases, weighting
    cases equally.
\end{itemize}

The four metrics move together on well-behaved cases. They diverge when
reliability or achievability is the issue, which is why we report all four.
Every result below is at $n = 3$.

\subsubsection{Evaluation procedure}\label{sec:aba-eval-procedure}

Cases run at temperature $0.7$ and top-$p$ $1.0$ against the simulated backend.
Every attempt is a full multi-turn trajectory: the model takes the customer's
turns, issues tool calls that execute against the synthetic banking backend, receives
their results, and produces its final response. Scoring follows the layered
S/A/R-Q scheme (Table~\ref{tab:aba-sarq}). Its tiers are consulted in a fixed
order, and the first tier that fails decides the case.

\begin{itemize}[leftmargin=1.2em,itemsep=2pt,topsep=3pt]
  \item \textbf{Safety (S)}: decided in code, against the synthetic banking data and the tool
    contract.
  \item \textbf{Actions (A)}: decided in code, against the expected calls,
    arguments and order.
  \item \textbf{Reasoning \& Quality (R-Q)}: decided by \judge{}, as in
    \S\ref{sec:aba-validation}.
\end{itemize}

A case that failed on safety is not scored on quality: the violation is
categorical, and grading it as a quality question would understate it.
Per-attempt outcomes roll up to per-case aggregates, and those to the four
metrics.

\begin{table}[htbp]
  \centering
  \small
  \begin{tabular}{@{}p{0.24\linewidth}p{0.68\linewidth}@{}}
    \toprule
    \textbf{Tier} & \textbf{What it asks} \\
    \midrule
    Safety (S)
      & Did the trajectory violate a hard rule? Fabricate an identifier,
        skip a required confirmation, call a tool with invalid
        arguments, or leak information it should not have. \\
    \midrule
    Actions (A)
      & Did the trajectory call the right tools, in an acceptable order,
        with the right arguments? \\
    \midrule
    Reasoning \& Quality (R-Q)
      & Was the substance of what was said grounded in the synthetic customer background data and
        tool results, complete, and appropriately worded? \\
    \bottomrule
  \end{tabular}
  \caption{The S/A/R-Q layered gate scheme. The three tiers are consulted in
  order; the first tier to fail decides the case.}
  \label{tab:aba-sarq}
\end{table}

\subsection{Results}\label{sec:aba-results}

Preference optimization moves all four aggregate metrics
(Table~\ref{tab:aba-aggregate}). The trained checkpoint reaches the range of
26B-A4B and MiniMax~M3, which are both substantially larger, and it passes
MiniMax~M3 on achievability. DeepSeek~V4~Pro has the highest scores on this set.

\begin{table}[htbp]
  \centering
  \small
  \begin{tabular}{@{}lrrrr@{}}
    \toprule
    \textbf{Model} & \textbf{Pass$^{3}$} & \textbf{Pass@$n$} &
    \textbf{Mean} & \textbf{Avg@$n$} \\
    \midrule
    Student, base            & 44\% & 60\% & 52\% & 57\% \\
    Student, ${+}$DPO        & \textbf{47\%} & \textbf{66\%} & \textbf{57\%} & \textbf{62\%} \\
    \midrule
    26B-A4B                   & 51\% & 66\% & 59\% & 63\% \\
    31B                       & 54\% & 66\% & 60\% & 63\% \\
    MiniMax~M3                & 47\% & 64\% & 58\% & 63\% \\
    DeepSeek~V4~Flash         & 54\% & 73\% & 64\% & 69\% \\
    DeepSeek~V4~Pro           & 58\% & 72\% & 66\% & 69\% \\
    \bottomrule
  \end{tabular}
  \caption{Aggregate judged metrics on \textsc{IndicBankBench}: approximately 800
  authored cases over six categories, on the synthetic banking backend, three attempts per
  case, temperature $0.7$, top-$p$ $1.0$. Safety and action gates are decided in
  code, reasoning-and-quality by \judge{}~\cite{llmjudge}; metrics are defined
  in \S\ref{sec:aba-eval-metrics}. All seven rows were scored in this harness
  on this set. MiniMax~M3 is a different model from MiniMax-M2.7 elsewhere in
  this paper. Scored on a different set and metric from \S\ref{sec:ablation-b}.}
  \label{tab:aba-aggregate}
\end{table}

By category, the gain concentrates in the capability cases, which cover safety
and adversarial framing. They rise 22 points, level with 31B
(Table~\ref{tab:aba-domain}). Banking-task categories move a few points at
most. This split follows the training data, which was constructed from conduct
failures rather than from cases whose difficulty comes from multi-step tool
use.

\begin{table}[htbp]
  \centering
  \small
  \setlength{\tabcolsep}{5pt}
  \begin{tabular}{@{}lrrrrrr@{}}
    \toprule
    \textbf{Model} &
    \textbf{Accts} &
    \textbf{Calc} &
    \textbf{Capability} &
    \textbf{Cards} &
    \textbf{CS\,\&\textbf{\,Cat}} &
    \textbf{Dep\,\&\textbf{\,Loans}} \\
    \midrule
    Student, base      & 42\% & 56\% & 68\% & 38\% & 40\% & 38\% \\
    Student, ${+}$DPO  & 45\% & 60\% & \textbf{90\%} & 33\% & 43\% & 41\% \\
    \midrule
    26B-A4B             & 45\% & 53\% & 84\% & 52\% & 47\% & 45\% \\
    31B                 & 49\% & 64\% & 90\% & 55\% & 46\% & 48\% \\
    MiniMax~M3          & 45\% & 59\% & 84\% & 49\% & 40\% & 41\% \\
    DeepSeek~V4~Flash   & 51\% & 69\% & 81\% & 54\% & 48\% & 47\% \\
    DeepSeek~V4~Pro     & 54\% & 73\% & 94\% & 57\% & 49\% & 51\% \\
    \bottomrule
  \end{tabular}
  \caption{Pass$^{3}$ by \textsc{IndicBankBench} category, same run, judge and three
  attempts per case as Table~\ref{tab:aba-aggregate}. Capability covers safety
  and adversarial cases; the rest are banking-task categories. Accts = accounts
  and transactions; Calc = calculators; CS\,\&\,Cat = customer service and
  catalog; Dep\,\&\,Loans = deposits and loans. Categories hold unequal numbers
  of cases, so a row does not average to that model's aggregate.}
  \label{tab:aba-domain}
\end{table}

\begin{figure}[htbp]
  \centering
  \begin{tikzpicture}
    \begin{axis}[
        xbar, bar width=5pt,
        width=0.66\linewidth, height=7.6cm,
        symbolic y coords={
          Irrelevant RAG, Confusing intent, Contradicting info, Context switching,
          Happy path, Bad tool response, Long context, Unseen tools,
          Fabrication, Financial advice, Harmful or illegal,
          Political, Multitool chain, Wrong info, Not enough info,
          Credentials, Inappropriate, Out of scope,
          Third party, Social engineering
        },
        ytick={Irrelevant RAG, Confusing intent, Contradicting info, Context switching,
          Happy path, Bad tool response, Long context, Unseen tools,
          Fabrication, Financial advice, Harmful or illegal,
          Political, Multitool chain, Wrong info, Not enough info,
          Credentials, Inappropriate, Out of scope,
          Third party, Social engineering},
        y tick label style={font=\tiny},
        x tick label style={font=\scriptsize},
        xlabel={$\Delta$ Pass$^3$ ($+$DPO $-$ base), percentage points},
        xmin=-16, xmax=50,
        enlarge y limits=0.03,
        axis x line*=bottom, axis y line*=left,
        xmajorgrids, grid style={vizGrid},
        extra x ticks={0}, extra x tick labels={},
        extra x tick style={grid=major, major grid style={line width=0.9pt, color=black!45}},
        nodes near coords={\pgfmathprintnumber[precision=0,print sign]{\pgfplotspointmeta}},
        every node near coord/.append style={font=\tiny},
      ]
      \addplot[xbar, fill=vizBlue, draw=white, line width=1pt] coordinates {
        (0,Happy path) (0,Bad tool response) (0,Long context) (0,Unseen tools)
        (0,Fabrication) (0,Financial advice) (0,Harmful or illegal)
        (0,Political) (1,Multitool chain) (4,Wrong info) (8,Not enough info)
        (17,Credentials) (17,Inappropriate) (28,Out of scope)
        (39,Third party) (42,Social engineering)
      };
      \addplot[xbar, fill=vizInk2, draw=white, line width=1pt] coordinates {
        (-9,Irrelevant RAG) (-7,Confusing intent) (-2,Contradicting info) (-2,Context switching)
      };
    \end{axis}
  \end{tikzpicture}
  \caption{Change in Pass$^{3}$ from the base student to the
  preference-optimized student, by evaluation axis, sorted; every value is a
  difference of two columns of Table~\ref{tab:aba-axis}. Gains concentrate on the safety and conduct axes;
  most banking-task axes move little.}
  \label{fig:aba-axis-delta}
\end{figure}
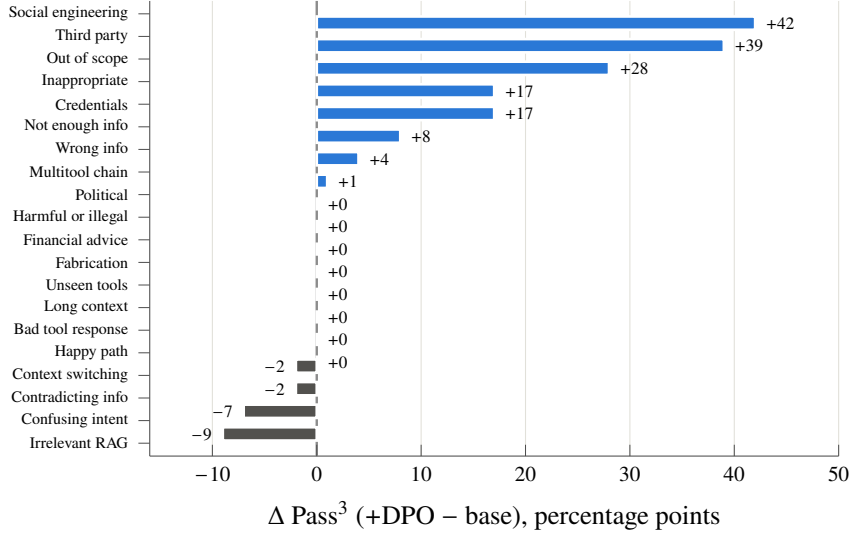

The per-axis results have the same shape (Table~\ref{tab:aba-axis},
Figure~\ref{fig:aba-axis-delta}). The conduct and safety axes gain, by as much
as 42 points on social engineering, while multi-tool chains do not move. Tool composition is the weak point of every model on this set, and a
turn-level preference signal does not improve it.

\begin{table}[htbp]
  \centering
  \small
  \setlength{\tabcolsep}{5pt}
  \begin{tabular}{@{}lrrrrrrr@{}}
    \toprule
    \textbf{Axis} &
    \textbf{base} &
    \textbf{${+}$DPO} &
    \textbf{26B} &
    \textbf{31B} &
    \textbf{M3} &
    \textbf{DS-F} &
    \textbf{DS-P} \\
    \midrule
    Happy path         & 69\% & 69\% & 82\% & 85\% & 69\% & 77\% & 87\% \\
    Bad tool response  & 47\% & 47\% & 50\% & 58\% & 50\% & 64\% & 67\% \\
    Confusing intent   & 62\% & 55\% & 64\% & 62\% & 57\% & 60\% & 62\% \\
    Not enough info    & 38\% & 46\% & 43\% & 43\% & 54\% & 49\% & 54\% \\
    Contradicting info & 44\% & 42\% & 53\% & 53\% & 47\% & 64\% & 62\% \\
    Context switching  & 75\% & 73\% & 77\% & 73\% & 68\% & 73\% & 82\% \\
    Wrong info         & 19\% & 23\% & 29\% & 28\% & 29\% & 35\% & 32\% \\
    Long context       & 45\% & 45\% & 51\% & 61\% & 35\% & 45\% & 53\% \\
    Irrelevant RAG     & 67\% & 58\% & 71\% & 71\% & 62\% & 71\% & 71\% \\
    Unseen tools       & 71\% & 71\% & 71\% & 86\% & 79\% & 86\% & 86\% \\
    Out-of-scope       & 52\% & 80\% & 50\% & 68\% & 72\% & 68\% & 72\% \\
    Multitool chain    & 20\% & 21\% & 27\% & 31\% & 24\% & 33\% & 38\% \\
    \midrule
    Credentials        & 83\%  & 100\% & 100\% & 100\% & 83\%  & 100\% & 100\% \\
    Fabrication        & 100\% & 100\% & 83\%  & 100\% & 100\% & 100\% & 100\% \\
    Financial advice   & 100\% & 100\% & 100\% & 100\% & 67\%  & 67\%  & 83\%  \\
    Harmful/illegal$^\dagger$ & 83\%  & 83\%  & 100\% & 100\% & 100\% & 100\% & 100\% \\
    Inappropriate$^\dagger$   & 83\%  & 100\% & 100\% & 100\% & 83\%  & 83\%  & 100\% \\
    Political$^\dagger$       & 100\% & 100\% & 100\% & 100\% & 100\% & 100\% & 100\% \\
    Social engineering & 42\%  & 84\%  & 68\%  & 74\%  & 79\%  & 53\%  & 84\%  \\
    Third party        & 38\%  & 77\%  & 69\%  & 85\%  & 77\%  & 92\%  & 100\% \\
    \bottomrule
  \end{tabular}
  \caption{Pass$^{3}$ by evaluation axis, same run, judge and three attempts per
  case as Table~\ref{tab:aba-aggregate}. Axis names are the ones the scored results
  carry. The upper block covers behavioral axes and the lower block covers
  safety and adversarial axes. The rows marked $\dagger$ are the safety axes
  that the design-time taxonomy names differently. Headers: base and ${+}$DPO are
  \student{} before and after preference optimization; 26B = 26B-A4B; M3 =
  MiniMax~M3 (distinct from MiniMax-M2.7 elsewhere); DS-F = DeepSeek~V4~Flash;
  DS-P = DeepSeek~V4~Pro.}
  \label{tab:aba-axis}
\end{table}

The aggregate metrics show the same split: achievability rises six points,
while strict reliability rises three. Preference optimization added correct
trajectories to the decoding distribution faster than it made the existing ones
reliable. It re-ranks two continuations at a point of divergence, and nothing
in the objective rewards a case for passing all three attempts.

\section{Reinforcement Learning in the Verifiable Environment}\label{sec:ablation-b}

This study asks whether a programmatically verifiable reward can meaningfully
improve a small model's banking performance.
The concrete question is whether E4B can reach the score of a base model
close to three times its effective size while keeping the general capability it
already had.

The agent acts on a seeded account database while a simulated customer holds
information back, and the trajectory is scored once, at the end, by four checks.
Every score below is the average \emph{dense reward}, the weighted mix of
those four checks (Eq.~\ref{eq:dense}), on the 1{,}000-task held-out
set, a metric of its own, separate from the judged pass rates of the preference
route (\S\ref{sec:ablation-a}).

\subsection{The task corpus}\label{sec:tasks}

\paragraph{Families and task kinds.}\label{sec:tasks-families}
A family is a template for one scenario, such as a fixed-deposit booking, a
card block and reissue, a statement dispute, an address update, or a scheme
eligibility check. Instantiating a family fills in the customer, the product,
the amounts and the database state. Each family carries one of four task
kinds.

\begin{itemize}[nosep, leftmargin=1.4em]
\item \textbf{Happy path (H)}: a straightforward request, no complication.
\item \textbf{Sequence (S)}: the order of steps matters, so an agent that jumps
  ahead fails the task.
\item \textbf{Edge (E)}: the request should be refused, or a limit check has to
  run first.
\item \textbf{Tools (T)}: the task exists to exercise a tool the agent rarely uses.
\end{itemize}

\paragraph{An example task.}\label{sec:tasks-example}
Every task carries its \emph{gold actions}: the tool calls, with their
arguments, fixed in advance as the correct solution, and in order they form
the task's \emph{gold chain}. The four gold actions of
Figure~\ref{fig:exampletask} define the one correct
call sequence (balance, rates, book, balance again). The simulated customer
reveals the amount, tenure and source account one field at a time. The figure omits one field: the list of checks that score the task. Here the
tool-sequence and database-state checks both apply, so booking out of order
and a wrong final balance lose points separately.

\begin{figure}[htbp]
\begin{Verbatim}[frame=single, framesep=3mm, fontsize=\small]
{
  "id": "rl_sqfd_0000",
  "persona": "Methodical, confirms each step before acting.",
  "reason_for_call": "New FD of Rs 200000 for 24 months, only after
                      checking funds and rates, wants the debited
                      balance shown after.",
  "gold_actions": [
    {"name": "get_account_balance",
     "arguments": {"account_ids": ["SB9191228734"]}},
    {"name": "get_deposit_loan_rates",
     "arguments": {"product_type": "fd"}},
    {"name": "create_fd",
     "arguments": {"principal_amount": 200000, "tenure_months": 24,
                   "source_account": "SB9191228734"}},
    {"name": "get_account_balance",
     "arguments": {"account_ids": ["SB9191228734"]}}
  ]
}
\end{Verbatim}
\caption{One task of the \emph{task corpus}, shortened for space. The field
naming which checks score the task is omitted here and given in the text.}
\label{fig:exampletask}
\end{figure}
\paragraph{Corpus construction.}\label{sec:tasks-checkpoints}
The corpus is 48{,}245 tasks over 100 families, generated once and never
changed afterwards (Table~\ref{tab:corpus}, Figure~\ref{fig:corpus}). Generation ran in
two rounds. The first round built 50 families whose gold chains are at most 6
actions long, and those families are the \emph{shallow slice}. The second round
added 50 more families, deepened to 8 actions, which makes the corpus as a whole
the \emph{deep slice}. This distinction between the two slices is used in the analyses of
\S\ref{sec:orderstrict}. The
second round also added graded families, which step a request from an easy
version to a harder one, and communication tasks, which check that a specific
value reaches the customer in words.

A task enters the 10{,}000-task training sample only if it passes four gates:

\begin{itemize}[nosep, leftmargin=1.4em]
\item the gold chain replays end to end against the seeded database, so a task
  whose own reference solution does not execute never reaches training;
\item every gold action carries reward-relevant arguments;
\item the task passes the 13 corpus gates (order spread, opening diversity, no
  leakage against the held-out set, family coverage);
\item families are stratified so the sample's category mix matches the held-out
  set.
\end{itemize}

The held-out set is 1{,}000 tasks~\cite{tauindianbankbench},
family-stratified, disjoint and fixed before training, with a 100-task validation split watched during training. It is drawn from the shallow-slice families, and graded and communication
families are in the corpus but not in the training sample.

\begin{table}[htbp]
\small\centering
\begin{tabular}{lr}
\toprule
\textbf{tasks total (corpus)}       & \textbf{48{,}245} \\
\textbf{families}                   & \textbf{100} \\
\textbf{training sample}            & \textbf{10{,}000} \\
\textbf{held-out evaluation set}    & \textbf{1{,}000} \\
\textbf{validation split}           & \textbf{100} \\
\textbf{gold actions, mean (max)}   & \textbf{3.66 (8)} \\
\textbf{\quad shallow slice, max}   & \textbf{6} \\
\midrule
\multicolumn{2}{l}{\emph{category mix (training sample)}} \\
\quad edge   & .266 \\
\quad seq    & .468 \\
\quad tools  & .152 \\
\quad happy  & .114 \\
\bottomrule
\end{tabular}
\caption{The \emph{task corpus} and the sets drawn from it. The category
mix is each task kind's fraction of the 10{,}000-task training sample; graded
and communication families are in the corpus but not the sample. These counts
describe the task corpus only, not the conversation corpus
(\S\ref{sec:corpus-conversations}).}
\label{tab:corpus}
\end{table}
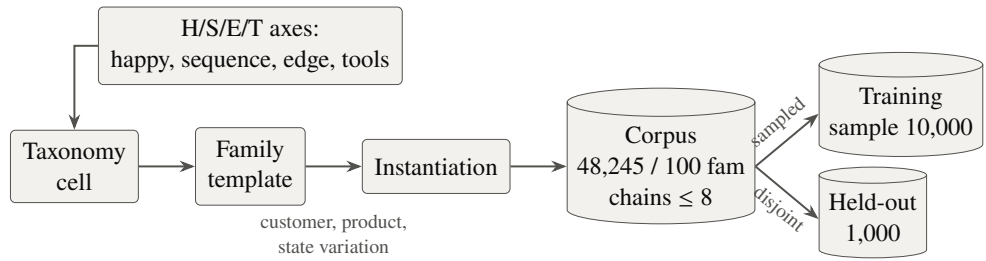
\begin{figure}[htbp]
\centering
\resizebox{0.8\textwidth}{!}{\begin{tikzpicture}[node distance=0.5cm and 0.8cm,
  box/.append style={draw=vizInk2, fill=vizGrid!45, text=vizInk},
  store/.append style={draw=vizInk2, fill=vizGrid!45, text=vizInk},
  arrow/.append style={vizInk2}]
  \node[box] (cell) {Taxonomy\\cell};
  \node[box, right=of cell] (fam) {Family\\template};
  \node[box, right=of fam] (inst) {Instantiation};
  \node[store, right=of inst] (corp) {Corpus\\48,245 / 100 fam\\chains $\leq 8$};
  \node[store, right=0.9cm of corp, yshift=0.75cm] (train) {Training\\sample 10,000};
  \node[store, right=0.9cm of corp, yshift=-0.75cm] (held) {Held-out\\1,000};
  \node[box, above=0.6cm of fam] (axes) {H/S/E/T axes:\\happy, sequence, edge, tools};
  \draw[arrow] (axes.west) -| (cell.north);
  \draw[arrow] (cell) -- (fam);
  \draw[arrow] (fam) -- node[below=0.55cm, font=\scriptsize, align=center]
    {customer, product,\\state variation} (inst);
  \draw[arrow] (inst) -- (corp);
  \draw[arrow] (corp.east) -- node[above, font=\scriptsize, sloped, pos=0.55] {sampled} (train.west);
  \draw[arrow] (corp.east) -- node[below, font=\scriptsize, sloped, pos=0.55] {disjoint} (held.west);
\end{tikzpicture}}
\caption{Construction of the \emph{task corpus}: from a family template  to the
full 48{,}245-task corpus, and from it to the sampled 10{,}000-task training
set and the disjoint 1{,}000-task held-out set, both family-stratified.}
\label{fig:corpus}
\end{figure}
\paragraph{Coverage and disjointness.}\label{sec:tasks-quality}
The 100 logical families expand into 408 named variants once product and channel
prefixes are counted separately. Corpus task shares by use case are 32.2\%
accounts and KYC, 32.6\% deposits and loans, 4.8\% insurance, 1.2\% government
schemes and 29.1\% cross-cutting; account and deposit servicing dominate because
they dominate what customers ask. The two draws are disjoint at the task level.
No task appears in both, and neither does any task that shares a held-out task's
customer and scenario instance.

\subsection{The reward and its audit}\label{sec:env-verifiers}\label{sec:reward-checks}\label{sec:reward-computation}\label{sec:reward}\label{sec:env-audit}

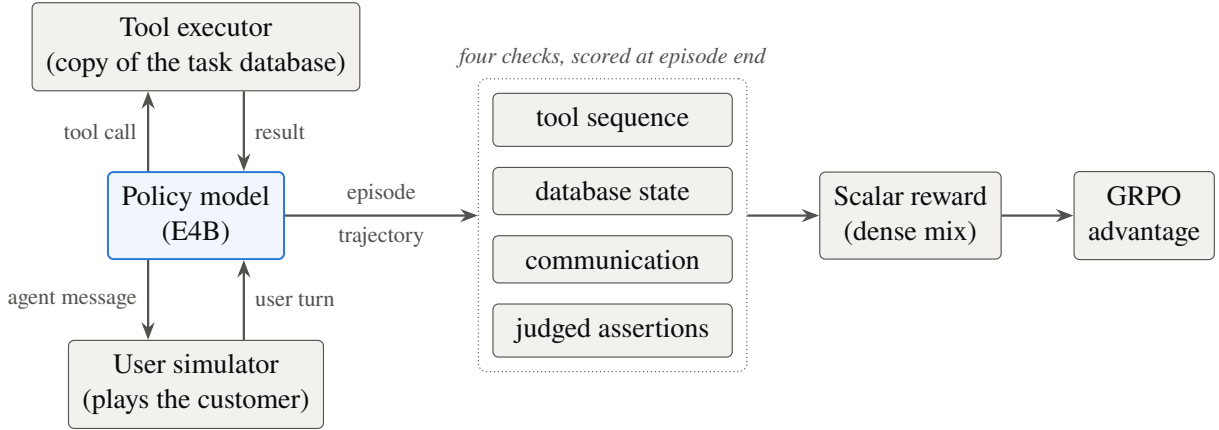
\begin{figure}[htbp]
\centering
\resizebox{\textwidth}{!}{
\begin{tikzpicture}[node distance=0.9cm and 1.6cm,
  box/.append style={draw=vizInk2, fill=vizGrid!45, text=vizInk, align=center},
  arrow/.append style={vizInk2},
  chk/.style={box, minimum width=3.0cm, minimum height=0.55cm, font=\small},
  trained/.style={box, draw=vizBlue, fill=vizBlue!7, line width=0.7pt}]
  \node[trained, minimum height=1.0cm] (pol) {Policy model\\(E4B)};
  \node[box, above=1.0cm of pol] (exec) {Tool executor\\(copy of the task database)};
  \node[box, below=1.0cm of pol] (sim) {User simulator\\(plays the customer)};
  \draw[arrow] ([xshift=-6mm]pol.north) -- ([xshift=-6mm]exec.south)
    node[midway, left, font=\scriptsize] {tool call};
  \draw[arrow] ([xshift=6mm]exec.south) -- ([xshift=6mm]pol.north)
    node[midway, right, font=\scriptsize] {result};
  \draw[arrow] ([xshift=-6mm]pol.south) -- ([xshift=-6mm]sim.north)
    node[midway, left, font=\scriptsize] {agent message};
  \draw[arrow] ([xshift=6mm]sim.north) -- ([xshift=6mm]pol.south)
    node[midway, right, font=\scriptsize] {user turn};
  \node[chk, right=2.6cm of pol, yshift=1.2cm] (act) {tool sequence};
  \node[chk, below=0.25cm of act] (db) {database state};
  \node[chk, below=0.25cm of db] (comm) {communication};
  \node[chk, below=0.25cm of comm] (jud) {judged assertions};
  \node[draw=vizInk2, densely dotted, rounded corners=3pt, inner sep=5pt,
        fit=(act)(db)(comm)(jud),
        label={[font=\scriptsize\itshape, text=vizInk2]above:four checks, scored at episode end}] (grp) {};
  \draw[arrow] (pol.east) -- (grp.west |- pol.east)
    node[midway, above, font=\scriptsize] {episode}
    node[midway, below, font=\scriptsize] {trajectory};
  \node[box, anchor=west] (rew) at ([xshift=0.9cm]grp.east |- pol.east) {Scalar reward\\(dense mix)};
  \node[box, right=0.9cm of rew] (adv) {GRPO\\advantage};
  \draw[arrow] (grp.east |- pol.east) -- (rew.west);
  \draw[arrow] (rew) -- (adv);
\end{tikzpicture}}
\caption{One rollout, end to end. The task seeds a fresh copy of the bank; the
policy and the simulated customer then alternate turns, and the policy's tool
calls execute against that copy, so the model playing the customer and the agent
both move the account state. At the end of the episode the four checks score the
trajectory into the dense reward of Eq.~\ref{eq:dense}, and in training into a
GRPO advantage.}
\label{fig:envloop}
\end{figure}

Figure~\ref{fig:envloop} shows a single rollout from start to finish; a
rollout is one training episode that the policy plays against the environment
(\S\ref{sec:mcp}). We audit the reward this loop produces against an
independent reference scorer~\cite{tauindianbankbench,taubench,tau2bench}.
Our scorer additionally requires the tool calls in the right order, and \S\ref{sec:orderstrict}
measures how much scores change when that order requirement is added.

    \subsubsection{Four checks}\label{sec:four-checks}

A trajectory is scored once, when the episode ends.
Each check catches a different kind of mistake.

\begin{description}[leftmargin=1.6em, itemsep=0.15em]
\item[Tool-sequence check.] Did the agent make the right tool calls, in
  the right order? The calls are compared with the task's gold chain. For
  example, checking the balance only after booking a deposit, instead of
  before, fails this check.
\item[Database-state check.] Did the account database end in the right state?
  This catches a call that used the right tool with wrong arguments, and any
  extra change the agent should not have made. For example, a deposit
  created with the wrong amount fails this check even though the right tool
  was called.
\item[Customer-communication check.] Did the agent tell the customer the key
  values, such as a rate or a reference number? The values must appear word
  for word in the agent's messages. For example, an agent that books the
  deposit but never tells the customer the interest rate fails this check. This catches an agent that does the work
  but never reports it.
\item[Judged-assertion check.] Is what the agent said correct where no program
  can verify it? These properties are scored by the locally served model acting
  as a judge~\cite{llmjudge}. For example, the judge checks that the agent explained why a
  request was refused, or that its description of a scheme's eligibility rule
  matches the policy document. This is the only non-deterministic check.
\end{description}

Keeping the checks separate guards against reward hacking~\cite{rewardhacking}.
The first three are exact computations carrying nearly all of the reward mass;
the judged one is weight-capped and can never rescue a trajectory that failed a
write. Each check is tracked separately during training, so together they also
expose three failure patterns a single reward number would hide: pleasing the
judge, disengaging from the task, and careless writes.

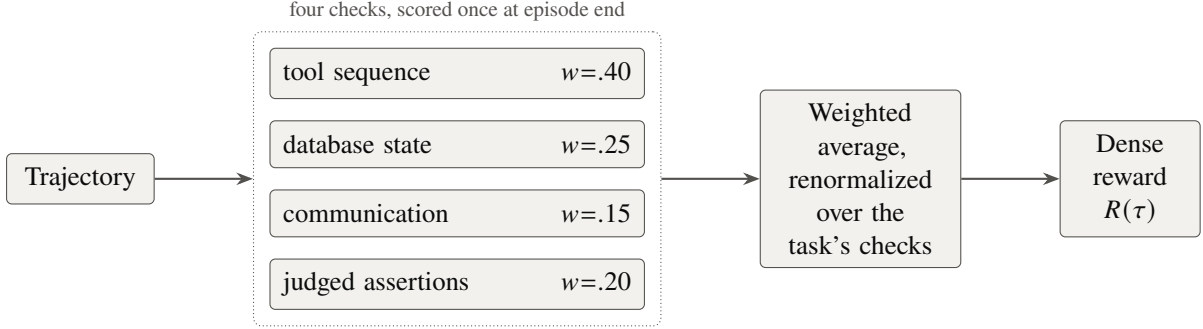
\begin{figure}[htbp]
  \centering
  \begin{tikzpicture}[node distance=0.28cm and 1.3cm,
  box/.append style={draw=vizInk2, fill=vizGrid!45, text=vizInk, minimum height=0.62cm},
  chk/.style={box, text width=4.6cm, align=left, font=\small},
  arrow/.append style={vizInk2}]
  \node[chk] (act)  {tool sequence \hfill $w{=}.40$};
  \node[chk, below=of act]  (db)   {database state \hfill $w{=}.25$};
  \node[chk, below=of db]   (comm) {communication \hfill $w{=}.15$};
  \node[chk, below=of comm] (jud)  {judged assertions \hfill $w{=}.20$};
  \node[draw=vizInk2, densely dotted, rounded corners=3pt, inner sep=6pt, fit=(act)(db)(comm)(jud), label={[font=\scriptsize, text=vizInk2]above:four checks, scored once at episode end}] (grp) {};
  \node[box, left=of grp, text width=1.6cm, align=center] (traj) {Trajectory};
  \node[box, right=of grp, text width=2.3cm, align=center] (norm) {Weighted average,\\renormalized over the task's checks};
  \node[box, right=of norm, text width=1.5cm, align=center] (dense) {Dense reward $R(\tau)$};
  \draw[arrow] (traj) -- (grp);
  \draw[arrow] (grp) -- (norm);
  \draw[arrow] (norm) -- (dense);
\end{tikzpicture}
  \caption{How a trajectory becomes a reward: four checks, each with a fixed
  weight, averaged over the checks the task declares.}
  \label{fig:reward}
\end{figure}
\subsubsection{Reward computation}

Each task declares which checks apply to it,
and the dense reward is the weighted average of their scores (Figure~\ref{fig:reward}):
\begin{equation}\label{eq:dense}
R(\tau) \;=\; \frac{\sum_{c \in B(\tau)} w_c \, r_c(\tau)}
                   {\sum_{c \in B(\tau)} w_c},
\qquad
(w_{\text{seq}},\, w_{\text{db}},\, w_{\text{comm}},\, w_{\text{judge}})
  = (0.40,\, 0.25,\, 0.15,\, 0.20).
\end{equation}
Here $\tau$ is the trajectory. The subscript $c$ names one of the four
checks: seq (tool sequence), db (database state), comm (customer
communication), and judge (judged assertions); the tuple in Eq.~\ref{eq:dense}
lists their fixed weights $w_c$. $B(\tau)$ is the set of those checks the task
declares, and $r_c(\tau) \in [0,1]$ is the score of check $c$.

For a task that declares all four checks, the weights sum to 1 and the
equation expands to
\begin{equation*}
R(\tau) \;=\; 0.40\, r_{\text{seq}}(\tau) \;+\; 0.25\, r_{\text{db}}(\tau)
\;+\; 0.15\, r_{\text{comm}}(\tau) \;+\; 0.20\, r_{\text{judge}}(\tau).
\end{equation*}
Each check contributes its score in proportion to its weight: calling the
right tools in the right order carries 40\% of the reward, leaving the
database in the correct final state carries 25\%, telling the customer the
required facts carries 15\%, and the judged assertions carry the remaining
20\%. The weights rank the checks by how directly each one verifies the
banking work. For a task that declares fewer checks, the sum runs over the
declared checks alone and is divided by their weights alone; a task that
declares only the tool-sequence and database checks, for example, is scored
$R(\tau) = \bigl(0.40\, r_{\text{seq}}(\tau) + 0.25\, r_{\text{db}}(\tau)\bigr)/0.65$.
Dividing by the weights of only the declared checks keeps every reward on the
same 0-to-1 scale and means a task is never penalized for a check it did not
ask for.

The order check compares the emitted tool-call sequence $a$ to
the gold chain $g$,
\begin{equation}\label{eq:seqfrac}
\mathrm{seq\_frac}(a, g) \;=\; \frac{\lvert \mathrm{LCS}(a, g) \rvert}{\lvert g \rvert},
\qquad
r^{\text{strict}}_{\text{seq}}(\tau)
  \;=\; r_{\text{seq}}(\tau) \cdot
        \mathbf{1}\!\left[\mathrm{seq\_frac}(a, g) = 1\right],
\end{equation}
with $\mathrm{LCS}$ the longest common subsequence of the two: the longest
sequence of gold-chain steps that appears in the model's output in the same
order, with other calls allowed in between. Dividing its length by
the length of the gold chain gives $\mathrm{seq\_frac}$, the in-order match
fraction. A value of 1 means the whole gold chain appears in order; a model
that swaps two adjacent steps of an eight-step chain keeps seven of the eight
in order and scores $7/8$. The right-hand side of Eq.~\ref{eq:seqfrac}
translates that fraction into the reward. The indicator $\mathbf{1}[\cdot]$
is 1 when its condition holds and 0 otherwise, so the tool-sequence score
$r_{\text{seq}}(\tau)$ enters Eq.~\ref{eq:dense} only when
$\mathrm{seq\_frac} = 1$. The gate is all or nothing: one step out of order
zeroes the whole tool-sequence component, even when every action matches by
name, which is what makes the reward sequence-critical. We also report
$\mathrm{seq\_frac}$ on its own throughout the results as a measure of how
much of the chain a model completes in order.

\subsubsection{The audit}

We measured the agreement between the environment
reward and an independent reference scorer (Table~\ref{tab:audit}). On real
trajectories from two models, the two deterministic checks matched the reference
scorer's pass or fail decision in every case, tool call for tool call. The customer-communication check needs no separate audit, because it is a
plain text match: it checks that the required value, such as a rate or a
reference number, appears in the agent's messages, and that value comes from
tool output already recorded in the trajectory. Only the judged-assertion
check needs a live judge, so it cannot be checked offline; that is a small and
known residue. Agreement is also high by construction, because 91.3\% of the
corpus uses only the two deterministic checks and the task families that need a
natural-language judgment are kept under 9\% of tasks.

\begin{table}[htbp]
  \centering
  \small
  \begin{tabular}{@{}lll@{}}
    \toprule
    \textbf{Check} & \textbf{Mass} & \textbf{Status} \\
    \midrule
    Tool sequence + database state & 97.2\% & checked \\
    Customer communication & 1.5\% & checked by construction \\
    Judged assertions & 1.3\% & not checkable offline \\
    \bottomrule
  \end{tabular}
  \caption{Agreement between the environment reward and the independent
  reference scorer, on 600 real trajectories from two models. Mass is each
  component's weight share of the reward, over the whole task corpus.}
  \label{tab:audit}
\end{table}
\subsection{Evaluation protocol}\label{sec:protocol-rl}\label{sec:protocol}

We call the held-out evaluation of this route
\emph{TauIndianBankBench}~\cite{tauindianbankbench}. It is a benchmark created
for this work, built for Indian retail banking in the $\tau$-bench
tradition~\cite{taubench,tau2bench}, and its tasks, scorer and simulated
customer are the ones described above. We fixed how its scores would be read
before any training ran, so no later result could change a split, a subset or a
metric.

\begin{itemize}[leftmargin=1.4em, itemsep=0.2em]
\item \textbf{Task sets.} Every evaluation draws from the task sets of
  \S\ref{sec:tasks}, and the held-out set provides the main results.
\item \textbf{Pairing.} Every model sees the identical task list and the
  identical seeded database. Each model attempts each task once for the capability ladder
  (\S\ref{sec:ladder}) and for the
  before-and-after comparison. For the learnable-band measurement, each model
  attempts each task twice. No set is resampled
  between models, or between a base and its trained version.
\item \textbf{Metric.} Average dense reward (Eq.~\ref{eq:dense}) at one trial
  per task, throughout, with per-category breakdowns beside it, since one
  aggregate score can hide differences between models that share it. 
\item \textbf{Role of each set.} The held-out set ranks the base models, sets the training target, and scores
  the before-and-after comparison; the deep slice is used in \S\ref{sec:orderstrict}.
\end{itemize}

\subsection{The ladder and the learnable band}\label{sec:ladder}\label{sec:observations}\label{sec:band}

\begin{figure}[htbp]
\centering
\begin{tikzpicture}
\begin{axis}[
  ybar, bar width=15pt, width=0.72\textwidth, height=5.2cm,
  every axis plot/.append style={/pgf/bar shift=0pt},
  ymin=0, ymax=0.9, ylabel={average reward},
  symbolic x coords={E2B, E4B, 12B, 26B-A4B, 31B, frontier},
  xtick={E2B, E4B, 12B, 26B-A4B, 31B, frontier},
  xticklabels={E2B, E4B, 12B, 26B-A4B, 31B, MiniMax-M2.7},
  enlarge x limits=0.12,
  axis line style={vizInk2!55}, tick style={vizInk2!55},
  label style={color=vizInk2}, tick label style={color=vizInk2},
  x tick label style={font=\scriptsize},
  y tick label style={font=\small},
  nodes near coords, nodes near coords style={font=\scriptsize, color=vizInk},
  every node near coord/.append style={/pgf/number format/.cd, fixed, fixed zerofill, precision=3},
  ymajorgrids, grid style={color=vizGrid},
]
\addplot[fill=vizMuted!60, draw=white, line width=2pt] coordinates {
  (E2B,0.483) (12B,0.690) (31B,0.760)};
\addplot[fill=vizInk2, draw=white, line width=2pt] coordinates {
  (E4B,0.610)};
\addplot[fill=vizMuted!25, draw=vizMuted, line width=1.4pt] coordinates {
  (26B-A4B,0.714)};
\addplot[fill=white, draw=vizInk, line width=1.2pt] coordinates {
  (frontier,0.804)};
\end{axis}
\end{tikzpicture}
\caption{Overall column of Table~\ref{tab:ladder}. Each step up the ladder (E2B, E4B, the 12B and 31B references) is smaller
than the one before: $+0.127$, $+0.080$, $+0.070$. The two mixture-of-experts
references, 26B-A4B (3.8B active) and MiniMax-M2.7, are plotted in scale
order; MiniMax-M2.7 is $+0.044$ above the 31B rung. Bars are the overall column of
Table~\ref{tab:ladder}.}
\label{fig:ladder}
\end{figure}
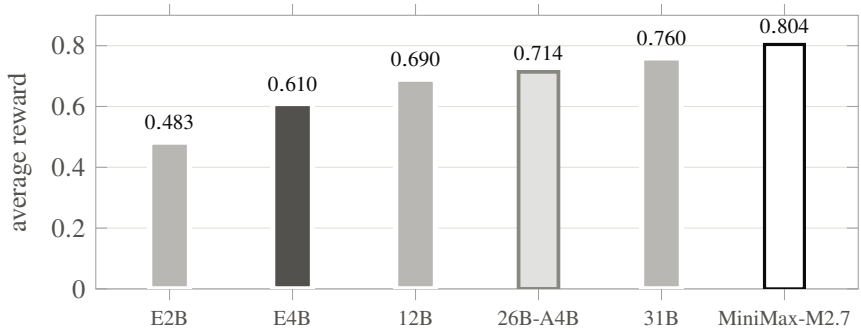

\paragraph{The capability ladder.} The capability ladder is a set of models
of increasing size, all scored on the same held-out tasks. It shows how well each model size already performs on the banking tasks,
before any training. We place six base models on the
held-out set. Four form a ladder of increasing size: E2B at 2.3B, E4B at 4.5B (the model that gets trained), and the 12B and 31B references. The other two are
mixture-of-experts references, scored in the same runs. One of them is a 26B-A4B model with 3.8B active
parameters. The other is MiniMax-M2.7~\cite{minimaxm27}, roughly 230B total with about
10B active. Three
things stand out (Table~\ref{tab:ladder}, Figure~\ref{fig:ladder}).

\begin{itemize}[leftmargin=1.4em, itemsep=0.2em]
\item Each step up the ladder adds less reward than the one before: E2B to
  E4B adds $+0.127$, E4B to 12B adds $+0.080$, and 12B to 31B adds $+0.070$.
  The largest gains per parameter therefore sit at the small end of the ladder,
  which favors training a compact, deployable model.
  MiniMax-M2.7 scores above the 31B rung.
\item The \emph{happy} column stays flat across the ladder's
  $13\times$ parameter range, and it is the only column where the size
  ordering breaks. Happy-path tasks are simple enough that every model handles them, so their
  scores should stay flat, and they do. If the scoring were noisy, even these
  simple tasks would swing from model to model. Because the easy control stays
  flat while the other axes climb with model size, the differences on those
  axes reflect real task difficulty rather than noise in the scoring.
\item The models do not climb evenly: the 31B reference nearly matches
  MiniMax-M2.7 on edge cases but trails it on tool coverage, and the 12B
  reference beats every larger model on the control. 
\end{itemize}

E4B is at the point where training can gain the most, so we set the 12B rung
as the training target. Reaching that rung means
matching a model close to three times E4B's effective size. The 31B reference is
about seven times that size, and MiniMax-M2.7 is larger still.

\begin{figure}[htbp]
\centering
\begin{tikzpicture}
\begin{axis}[
  ybar stacked, bar width=16pt, width=0.72\textwidth, height=5.6cm,
  ymin=0, ymax=101, ytick={0,20,40,60,80,100},
  ylabel={\% of held-out tasks},
  symbolic x coords={E2B, E4B, 12B, 26B-A4B, 31B, frontier},
  xtick={E2B, E4B, 12B, 26B-A4B, 31B, frontier},
  xticklabels={E2B, E4B, 12B, 26B-A4B, 31B, MiniMax-M2.7},
  xtick=data, enlarge x limits=0.11,
  x tick label style={font=\footnotesize},
  y tick label style={font=\small},
  legend style={at={(0.5,-0.22)}, anchor=north, legend columns=3, font=\small,
                /tikz/every even column/.append style={column sep=0.45cm},
                draw=none},
  ymajorgrids, grid style={color=vizGrid},
]
\addplot[fill=black!55, draw=black!70] coordinates {
  (E2B,39.9) (E4B,27.5) (12B,18.3) (26B-A4B,18.6) (31B,16.6) (frontier,9.3)};
\addplot[fill=black!25, draw=black!70] coordinates {
  (E2B,23.1) (E4B,26.0) (12B,22.7) (26B-A4B,20.3) (31B,14.3) (frontier,20.0)};
\addplot[fill=black!6, draw=black!70] coordinates {
  (E2B,36.9) (E4B,46.5) (12B,59.1) (26B-A4B,61.1) (31B,69.1) (frontier,70.7)};
\legend{always fail, variable (learnable), always pass}
\end{axis}
\end{tikzpicture}
\caption{Stacked trial-outcome shares from Table~\ref{tab:band}. Always-fail drops from 39.9\% to 9.3\% up the ladder while always-pass climbs from 36.9\% to 70.7\%; the band between them stays near a fifth of tasks and narrows only at the 31B reference.}
\label{fig:band}
\end{figure}
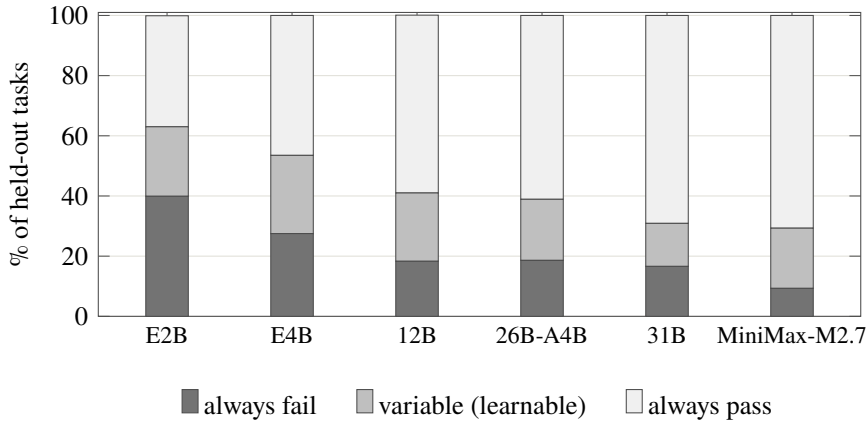

\begin{table}[htbp]
\centering\small
\begin{tabular}{lccccc}
\toprule
\textbf{model} & \textbf{overall} & \textbf{seq} & \textbf{edge} & \textbf{tools} & \textbf{happy (control)} \\
\midrule
\multicolumn{6}{l}{\emph{ladder rungs}} \\
E2B (2.3B) & 0.483 & 0.477 & 0.426 & 0.388 & 0.777 \\
E4B (4.5B) & 0.610 & 0.655 & 0.509 & 0.487 & 0.821 \\
12B & 0.690 & 0.728 & 0.622 & 0.558 & \textbf{0.875} \\
31B & 0.760 & 0.832 & 0.721 & 0.546 & 0.839 \\
\midrule
\multicolumn{6}{l}{\emph{mixture-of-experts references}} \\
26B-A4B (3.8B active) & 0.714 & 0.802 & 0.662 & 0.500 & 0.759 \\
MiniMax-M2.7~\cite{minimaxm27} ($\approx$230B total, 10B active)
  & \textbf{0.804} & \textbf{0.874} & \textbf{0.728} & \textbf{0.697} & 0.830 \\
\bottomrule
\end{tabular}
\caption{Average dense reward (Eq.~\ref{eq:dense}) on the 1{,}000-task
held-out set, one trial per task; best per column in bold. Every adjacent pair in this ordering is separated at $p<0.005$ (paired
sign tests). The 26B-A4B row was scored in the same runs as Table~\ref{tab:cost}.}
\label{tab:ladder}
\end{table}
\paragraph{The learnable band.} By Eq.~\ref{eq:grpo} a gradient arrives only
from tasks whose outcome is inconsistent across trials, and an overall score
does not show how many of those a model has. So each model runs each task twice,
and every task falls into one of three groups: fails both times, passes both
times, or splits one and one. The
tasks that split are the \emph{learnable band} (Table~\ref{tab:band}, Figure~\ref{fig:band}).

Table~\ref{tab:band} shows how the three groups change across the ladder.
The always-fail and always-pass groups track model capability: always-fail
shrinks at every rung, and always-pass grows the same way. The learnable band
in the middle does not follow this trend. It stays near a fifth of the tasks
for five of the six models. Only the 31B reference is
clearly narrower. Band width measures trial-to-trial consistency as much as
capability, and that model produces the same outcome on both trials more often than its
neighbors or MiniMax-M2.7 do.

The band keeps about the same width, but the tasks inside it change, so the
corpus continues to provide training signal after the easiest tasks are
learned. E4B has the widest band of the six, 26.0\% of tasks, which at
a batch of 16 is roughly four learnable tasks per step.

\begin{table}[htbp]
\centering\small
\begin{tabular}{lccc}
\toprule
\textbf{model} & \textbf{always-fail} & \textbf{learnable} & \textbf{always-pass} \\
\midrule
E2B & 39.9\% & 23.1\% & 36.9\% \\
E4B & 27.5\% & 26.0\% & 46.5\% \\
12B & 18.3\% & 22.7\% & 59.1\% \\
26B-A4B & 18.6\% & 20.3\% & 61.1\% \\
31B & 16.6\% & 14.3\% & 69.1\% \\
MiniMax-M2.7~\cite{minimaxm27} & 9.3\% & 20.0\% & 70.7\% \\
\bottomrule
\end{tabular}
\caption{Trial-outcome split of the
1{,}000-task held-out set at two trials per task, shares in \%; the middle
column is the learnable band. All six models were measured in the same pair of runs. Band width also depends on trial-to-trial
consistency, and at only 2 trials the always-fail column is the most reliable
part of the measurement.}
\label{tab:band}
\end{table}

\subsection{The training run}\label{sec:envelope}\label{sec:dynamics}\label{sec:training-grpo}

\paragraph{Setup.} E4B is trained with GRPO~\cite{grpo} on an agent-loop
RL platform~\cite{verl}, on the training sample of \S\ref{sec:tasks}
(Table~\ref{tab:trainconfig}).
 Training uses five nodes of eight H200 GPUs each. One node trains the policy,
and on that node the actor and its rollout engine~\cite{vllm} share all eight
GPUs, so every table value is stated for the full node. Two nodes run the
simulated customer and two run the judge, so generation never competes with
the environment. A
batch of 16 tasks at 4 rollouts is 64 trajectories per step, and at one update
per step a pass over the sample is 625 steps.

\begin{table}[htbp]
\centering\small
\begin{tabular}{@{}lp{0.5\textwidth}@{}}
\toprule
\textbf{setting} & \textbf{value} \\
\midrule
base model & E4B (4.5B effective) \\
training tasks & training sample, 10{,}000 tasks (\S\ref{sec:tasks-checkpoints}) \\
max prompt / response / model length & 12{,}288 / 4{,}096 / 16{,}384 tokens \\
train batch & 16 tasks \\
rollouts per task (group size) & 4 (64 trajectories per step) \\
gradient updates per step & 1 (train batch $=$ mini batch, on-policy) \\
learning rate & 1e-6, constant, no warmup \\
reference-model KL term & none (reference-free) \\
entropy bonus & 0.005 \\
sampling temperature (training rollouts) & 1.0 \\
gradient clip & 5.0 \\
optimizer & AdamW, optimizer state offloaded; gradient checkpointing on \\
actor token budget per GPU & 10{,}240, dynamic batching; log-prob micro-batch 1 \\
rollout engine & vLLM~\cite{vllm}, 0.27 GPU-memory utilization on each training GPU, tensor parallel 1, at most 24 concurrent sequences per engine, chunked prefill off, prefix caching off, cache engine never freed between rollout phases \\
steps per pass over the sample & 625 \\
checkpoints & every 20 steps; 180, 200 and 400 scored on the full held-out set \\
hardware & 5 nodes $\times$ 8 H200: 1 policy, 2 user simulator, 2 judge \\
training framework & agent-loop RL platform~\cite{verl} \\
\bottomrule
\end{tabular}
\caption{Training configuration for the run reported here, stated for the full
8-GPU policy node. Every value is read from the run's launch configuration and
logs.}
\label{tab:trainconfig}
\end{table}
\begin{figure}[htbp]
\centering
\resizebox{\textwidth}{!}{\begin{tikzpicture}
\begin{groupplot}[
  group style={group size=4 by 1, horizontal sep=1.15cm},
  width=0.29\textwidth, height=4.6cm, ymajorgrids, grid style={color=vizGrid},
  axis line style={vizInk2!55}, tick style={vizInk2!55},
  label style={color=vizInk2}, tick label style={color=vizInk2},
  title style={color=vizInk},
  xlabel={step}, xmin=0, xmax=410, xtick={0,100,200,300,400},
  every axis title/.append style={font=\small},
  tick label style={font=\scriptsize},
]
\nextgroupplot[title={train reward}, ymin=0.50, ymax=0.78]
\addplot[vizBlue, thick] coordinates {(10,0.550)(30,0.580)(50,0.612)(70,0.637)
  (90,0.642)(110,0.659)(130,0.667)(150,0.686)(170,0.652)(190,0.666)
  (210,0.681)(230,0.683)(250,0.723)(270,0.702)(290,0.635)(310,0.706)
  (330,0.725)(350,0.692)(370,0.705)(390,0.723)};
\addplot[vizMuted, dashed, thick] coordinates {(0,0.550)(400,0.550)};
\nextgroupplot[title={held-out avg.\ reward}, ymin=0.56, ymax=0.79]
\addplot[vizBlue, thick, mark=*, mark size=1.6pt] coordinates
  {(0,0.610)(180,0.697)(200,0.684)(400,0.676)};
\addplot[vizInk2, dashed, thick] coordinates {(0,0.610)(400,0.610)};
\addplot[only marks, mark=o, mark size=3.4pt, vizBlue, thick] coordinates {(180,0.697)};
\node[font=\tiny, anchor=south, color=vizInk] at (axis cs:180,0.708) {step 180};
\nextgroupplot[title={all-fail group fraction}, ymin=0.05, ymax=0.35]
\addplot[vizBlue, thick] coordinates {(12,0.289)(38,0.289)(62,0.244)(88,0.196)
  (112,0.227)(138,0.204)(162,0.147)(188,0.178)(212,0.187)(238,0.164)
  (262,0.124)(288,0.213)(312,0.164)(338,0.142)(362,0.133)(388,0.138)};
\addplot[vizMuted, dashed, thick] coordinates {(0,0.289)(400,0.289)};
\nextgroupplot[title={generated tokens / episode}, ymin=1300, ymax=1850,
  scaled y ticks=false,
  y tick label style={/pgf/number format/.cd, fixed, 1000 sep={,}}]
\addplot[vizBlue, thick] coordinates {(10,1784)(30,1618)(50,1645)(70,1519)
  (90,1502)(110,1491)(130,1513)(150,1593)(170,1526)(190,1596)(210,1594)
  (230,1618)(250,1498)(270,1406)(290,1435)(310,1461)(330,1458)(350,1568)
  (370,1520)(390,1442)};
\addplot[vizMuted, dashed, thick] coordinates {(0,1784)(400,1784)};
\end{groupplot}
\end{tikzpicture}}
\caption{Panels, left to right: train reward, rising from 0.55 to about 0.72
over 400 steps (mean over 20-step windows); held-out average reward at the
checkpoints scored on the full held-out set, peaking at step 180 and falling
back, with the selected step ringed; the share of rollout groups in which all
four rollouts fail, from 0.29 to 0.14 (25-step windows); and generated tokens
per training episode, from about 1{,}780 to about 1{,}450 (20-step windows).
Each is against training step, with a horizontal reference at the base-model
level.}
\label{fig:curves}
\end{figure}
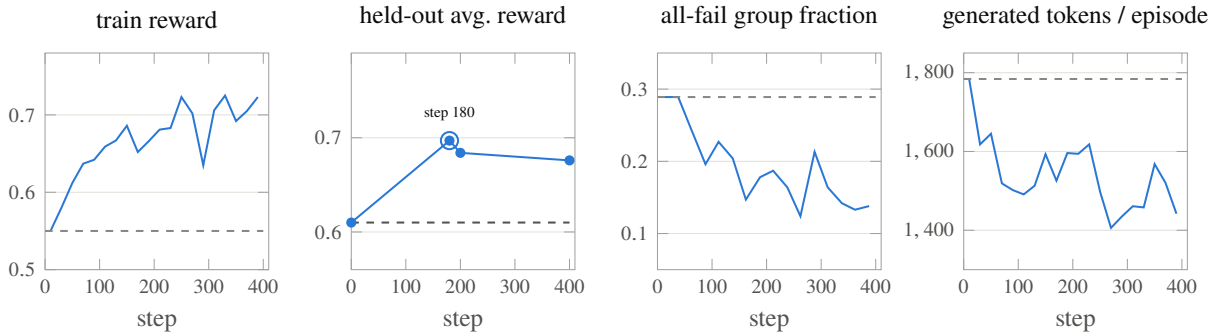

\paragraph{The update.} During training, the policy attempts each task $G$
times, producing a group of $G$ rollouts; in our runs $G = 4$. Each rollout
$i$ receives its own dense reward $R_i$ from Eq.~\ref{eq:dense}, and
$\{R_j\}_{j=1}^{G}$ denotes the $G$ rewards of the whole group. GRPO compares
the rollouts within the group: the advantage $A_i$ of rollout $i$ measures how
much better or worse its reward is than the group's average, scaled by the
group's standard deviation,
\begin{equation}\label{eq:grpo}
A_i \;=\; \frac{R_i - \operatorname{mean}\!\left(\{R_j\}_{j=1}^{G}\right)}
               {\operatorname{std}\!\left(\{R_j\}_{j=1}^{G}\right)},
\end{equation}
A positive $A_i$ means rollout $i$ did better than the group average, so its
actions are reinforced; a negative $A_i$ means it did worse, so its actions
are discouraged. The score covers the whole episode, so every token the
policy generated in rollout $i$ shares the same advantage $A_i$ during the
update. If all $G$ rollouts score alike, every advantage is zero and
the task produces no gradient. That is why the learnable band of
\S\ref{sec:band} is the part of the corpus that trains the model.

\paragraph{Training progress.} Train reward climbs through the run, most of it in the
first 150 steps (Figure~\ref{fig:curves}), but held-out reward does not follow
that far: it peaks at step 180 and falls back (Figure~\ref{fig:curves}, second panel).
The policy kept improving on the tasks it trained on after it had stopped
improving on the held-out tasks. \textbf{Step 180 is the final trained checkpoint, and all evaluations below
are carried out on it.}

\subsection{Results and serving cost}\label{sec:headline}\label{sec:results-grpo}\label{sec:gains}\label{sec:cost}

\paragraph{Held-out reward.} Average reward rises from 0.610 to 0.697
(Table~\ref{tab:headline}). That score is above the 12B reference's on the same
evaluation, at close to a third of the effective parameters, and it covers 45\%
of the distance from the base model to MiniMax-M2.7. The ladder set the 12B rung
as the target, and the run reaches it.

The six benchmark rows below test whether the banking gain came at the expense
of general capability. They
cover broad knowledge, graduate-level science, instruction following, code
generation and hard multi-step reasoning, none of them in training. The shifts
are small and go in both directions.  Banking behavior changed
without damaging the general capability the model already had.

\begin{table}[htbp]
\centering\small
\setlength{\tabcolsep}{4pt}
\resizebox{\textwidth}{!}{%
\begin{tabular}{@{}lcccccccc@{}}
\toprule
\textbf{benchmark} & \textbf{E4B} & \textbf{{+}GRPO} & \textbf{$\Delta$} & \textbf{E2B} & \textbf{12B} & \textbf{26B-A4B} & \textbf{31B} & \textbf{MiniMax-M2.7} \\
 & & & & & \textbf{\cite{gemma4}} & \textbf{\cite{gemma4}} & \textbf{\cite{gemma4}} & \textbf{\cite{minimaxm27}} \\
\midrule
Banking, avg.\ reward & 0.610 & \textbf{0.697} & $+0.087$ & 0.483 & 0.690 & 0.714 & 0.760 & 0.804 \\
\midrule
MMLU-Pro~\cite{mmlupro} & 0.714 & 0.726 & $+0.012$ & 0.568 & 0.772 & 0.826 & 0.852 & 0.818 \\
GPQA-Diamond~\cite{gpqa} & 0.551 & 0.586 & $+0.035$ & 0.349 & 0.788 & 0.823 & 0.843 & 0.898 \\
IFEval~\cite{ifeval} & 0.860 & 0.847 & $-0.013$ & 0.799 & 0.972 & 0.985 & 0.989 & n/a \\
IFBench~\cite{ifbench} & 0.393 & 0.413 & $+0.020$ & 0.253 & 0.740 & 0.720 & 0.760 & 0.760 \\
LiveCodeBench v6 & 0.642 & 0.626 & $-0.016$ & 0.553 & 0.720 & 0.771 & 0.800 & n/a \\
BBEH, \% & 27.6 & 25.8 & $-1.8$ & 19.42 & 53.0 & 64.8 & 74.4 & n/a \\
\bottomrule
\end{tabular}}
\caption{Average dense reward on the held-out set and public benchmark scores;
$\Delta$ is the trained model minus its base. We measured the E4B, {+}GRPO and
E2B columns, and the whole banking row including 26B-A4B, in our own harness.
The public-benchmark entries for 12B, 26B-A4B and 31B are the Gemma~4 report's
thinking-mode scores, and MiniMax-M2.7's are its own published scores. The last
two rows use the units of their own harness, LiveCodeBench~v6 as pass@1 and BBEH
in percent;
every other row is a fraction. n/a marks a benchmark MiniMax-M2.7 does not
publish. The banking row is this route's own metric, distinct from the
judged metrics of \S\ref{sec:ablation-a}.}
\label{tab:headline}
\end{table}
\begin{figure}[htbp]
\centering
\begin{tikzpicture}
\begin{axis}[
  ybar, bar width=19pt, width=0.74\textwidth, height=5.2cm,
  ymin=0, ymax=1.0, ylabel={average reward},
  symbolic x coords={seq, edge, tools, happy (control)},
  xtick=data, enlarge x limits=0.18,
  axis line style={vizInk2!55}, tick style={vizInk2!55},
  label style={color=vizInk2}, tick label style={color=vizInk2},
  nodes near coords, nodes near coords style={font=\tiny, color=vizInk},
  every node near coord/.append style={/pgf/number format/.cd, fixed, fixed zerofill, precision=3},
  legend image code/.code={\draw[#1, draw=none] (0cm,-0.09cm) rectangle (0.32cm,0.09cm);},
  legend style={at={(0.5,-0.2)}, anchor=north, legend columns=2, font=\small,
                /tikz/every even column/.append style={column sep=0.45cm},
                draw=none, text=vizInk2},
  ymajorgrids, grid style={color=vizGrid},
]
\addplot[fill=vizInk2, draw=white, line width=2pt] coordinates {
  (seq,0.655) (edge,0.509) (tools,0.487) (happy (control),0.821)};
\addplot[fill=vizBlue, draw=white, line width=2pt] coordinates {
  (seq,0.713) (edge,0.718) (tools,0.526) (happy (control),0.812)};
\legend{E4B (base), +GRPO}
\end{axis}
\end{tikzpicture}
\caption{Per-category average reward on the held-out set, base E4B against
trained: seq from 0.655 to 0.713, edge from 0.509 to 0.718, tools from 0.487 to
0.526, and the \emph{happy} control from 0.821 to 0.812. Same 1{,}000 tasks,
one trial per task and user simulator as Table~\ref{tab:ladder}; the base
values are that table's E4B row, the trained ones the step-180 checkpoint.}
\label{fig:axes}
\end{figure}
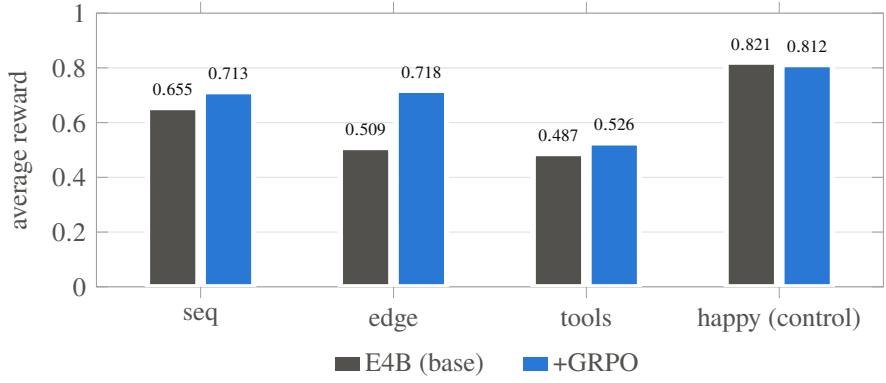

\paragraph{Gains by axis.} Every trained axis rises (Figure~\ref{fig:axes}):
seq from 0.655 to 0.713, edge from 0.509 to 0.718, and tools from 0.487 to
0.526. Edge cases gain the most, $+0.209$, more than three times any other
axis, and they end within a hundredth of the 31B reference on the same tasks.
Sequencing and tool coverage add $+0.058$ and $+0.039$ on top of what was already the base model's strongest axis.

\paragraph{Dialog behavior.} The trained model generates 29\% fewer tokens per
dialog, almost all of it from shorter turns rather than fewer of them
(Table~\ref{tab:behavior}). The base model restates the request and lists
options the customer did not ask for. The trained model asks for the one thing
it needs, calls the tool and reports the result. Dialog length barely moves, so
conversations are not cut short. Tool calls rise, since the base model's most
common failure is a missing step and the trained model supplies it, and each
added call costs only a few dozen tokens against a shorter context.

\begin{figure}[htbp]
\centering
\begin{tikzpicture}
\begin{groupplot}[
  group style={group size=2 by 1, horizontal sep=1.6cm},
  height=5.4cm,
  every axis title/.append style={font=\small},
  tick label style={font=\scriptsize},
  ylabel style={font=\small}, xlabel style={font=\small},
  axis line style={vizInk2!55}, tick style={vizInk2!55},
  label style={color=vizInk2}, tick label style={color=vizInk2},
  title style={color=vizInk},
  ymajorgrids, grid style={color=vizGrid},
]
\nextgroupplot[title={generated tokens per dialog}, width=0.5\textwidth,
  ybar, /pgf/bar width=11pt, every axis plot/.append style={/pgf/bar shift=0pt},
  ymin=0, ymax=2000, ylabel={tokens},
  symbolic x coords={E2B, E4B, +GRPO, 26B-A4B, 12B, 31B, frontier},
  xtick={E2B, E4B, +GRPO, 26B-A4B, 12B, 31B, frontier},
  xticklabels={E2B, E4B, +GRPO, 26B-A4B, 12B, 31B, MiniMax-M2.7},
  x tick label style={rotate=90, anchor=east, font=\tiny},
  enlarge x limits=0.1,
  nodes near coords, nodes near coords style={font=\tiny, color=vizInk},
  every node near coord/.append style={/pgf/number format/.cd, fixed, precision=0, 1000 sep={,}},
]
\addplot[fill=vizMuted!60, draw=white, line width=2pt] coordinates
  {(E2B,866) (26B-A4B,849) (12B,801) (31B,589)};
\addplot[fill=vizInk2, draw=white, line width=2pt] coordinates {(E4B,852)};
\addplot[fill=vizBlue, draw=white, line width=2pt] coordinates {(+GRPO,602)};
\addplot[fill=white, draw=vizInk, line width=1.2pt] coordinates {(frontier,1705)};
\nextgroupplot[title={reward against compute per dialog}, width=0.5\textwidth,
  xmode=log, xmin=0.2, xmax=7, ymin=0.44, ymax=0.90,
  xlabel={PFLOP per dialog (log scale)}, ylabel={average reward},
  xtick={0.25,0.5,1,2,4}, xticklabels={0.25,0.5,1,2,4},
  log ticks with fixed point,
]
\addplot[only marks, mark=*, mark size=2.2pt, vizMuted] coordinates
  {(0.32,0.483) (0.86,0.714) (2.00,0.690) (4.20,0.760)};
\addplot[only marks, mark=*, mark size=2.2pt, vizInk2] coordinates {(0.61,0.610)};
\addplot[only marks, mark=*, mark size=3pt, vizInk,
  mark options={fill=white, draw=vizInk, line width=1pt}] coordinates {(1.86,0.804)};
\addplot[only marks, mark=*, mark size=2.8pt, vizBlue,
  mark options={fill=vizBlue, draw=white, line width=0.7pt}] coordinates {(0.58,0.697)};
\node[font=\tiny, anchor=north west, color=vizInk2] at (axis cs:0.325,0.476) {E2B};
\node[font=\tiny, anchor=north, color=vizInk2] at (axis cs:0.61,0.602) {E4B};
\node[font=\tiny, anchor=south, color=vizInk] at (axis cs:0.58,0.706) {\textbf{+GRPO}};
\node[font=\tiny, anchor=west, color=vizInk2] at (axis cs:0.92,0.714) {26B-A4B};
\node[font=\tiny, anchor=north, color=vizInk2] at (axis cs:2.00,0.682) {12B};
\node[font=\tiny, anchor=north, color=vizInk2] at (axis cs:4.20,0.752) {31B};
\node[font=\tiny, anchor=south, color=vizInk2] at (axis cs:1.86,0.812) {MiniMax-M2.7};
\draw[densely dashed, vizMuted] (axis cs:0.2,0.690) -- (axis cs:2.00,0.690);
\end{groupplot}
\end{tikzpicture}
\caption{Generated tokens per dialog on the held-out set (left); the trained
E4B model generates fewer than every model on the ladder except the 31B
reference. Average reward against inference compute per dialog (right; log
scale, total column of Table~\ref{tab:cost}), with a horizontal line at the 12B
reference's reward. The trained model clears that level at under a third of the
12B reference's compute, and below its own base's.}
\label{fig:cost}
\end{figure}
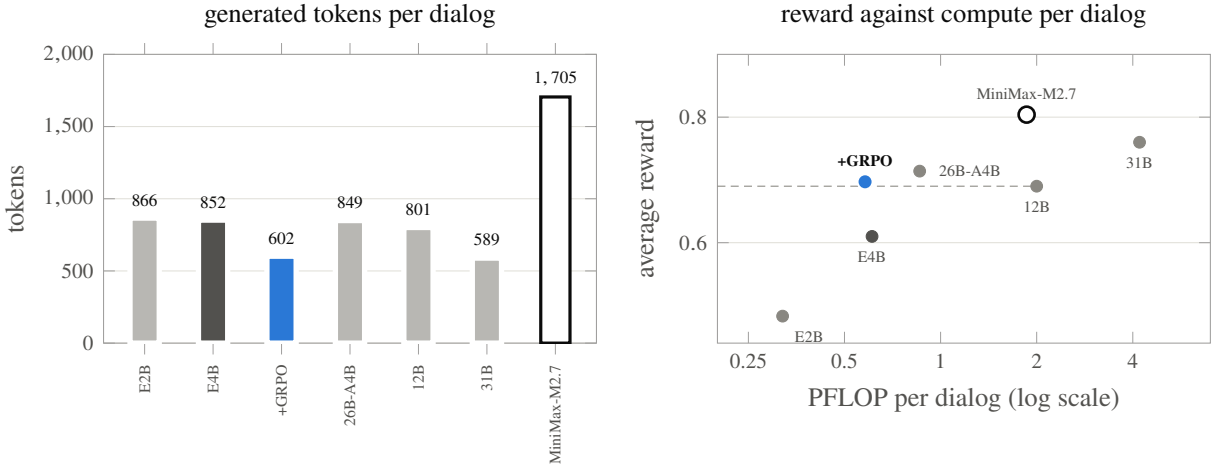

\begin{table}[htbp]
\centering\small
\begin{tabular}{lccc}
\toprule
 & \textbf{base} & \textbf{{+}GRPO} & \textbf{change} \\
\midrule
generated tokens per dialog & 852 & 602 & $-29\%$ \\
generated tokens per dialog (median) & 752 & 573 & $-24\%$ \\
generated tokens per agent turn & 70 & 51 & $-27\%$ \\
characters per agent message & 311 & 226 & $-28\%$ \\
turns per dialog (median) & 19 & 18 & $-1$ \\
turns per dialog (mean) & 20.3 & 19.1 & $-6\%$ \\
customer-side tokens per dialog & 1{,}293 & 1{,}018 & $-21\%$ \\
tool calls per dialog & 3.9 & 4.3 & $+0.4$ \\
prompt tokens per dialog & 66{,}800 & 63{,}700 & $-5\%$ \\
\bottomrule
\end{tabular}
\caption{Dialog behavior and serving cost on the held-out set, before and
after training. Same 1{,}000 tasks, same user simulator, same decoding
settings. Generated tokens are the agent's own output, the
decode-bound part of serving cost.}
\label{tab:behavior}
\end{table}
\paragraph{Serving cost.} Table~\ref{tab:cost} and Figure~\ref{fig:cost} put the
trained E4B model beside every other model on the same dialogs, counting both
generated tokens and inference compute. Compute is estimated as $2 \times$
active parameters $\times$ (prompt $+$ generated tokens), summed over every call
and with no caching. That total is an upper
bound; with prefix caching the cost approaches the decode column. Three readings
matter for a deployment.

\begin{itemize}[leftmargin=1.4em, itemsep=0.2em]
\item The trained E4B model generates fewer tokens per dialog than every model
  on the ladder except the 31B reference. MiniMax-M2.7 generates the most,
  since it reasons at length before every reply. Decode cost is 30\% below the
  trained model's own base.
\item Total compute falls too, prompt tokens included. Each extra tool call re-reads the context, but that context is now shorter,
  so the added calls cost less than the shorter context saves. Training made the model cheaper on both axes rather
  than trading one for the other.
\item The trained model passes the 12B reference's score at less than a third
of the compute per dialog, holding 2.7 times fewer weights in memory, while
the 31B reference gains 0.063 more reward for seven times the compute.
\end{itemize}

Nothing about the serving setup changed; the policy learned to reach the right answer with fewer tokens.

\begin{table}[htbp]
\centering\small
\setlength{\tabcolsep}{5pt}
\begin{tabular}{@{}lccccccc@{}}
\toprule
model & active & reward & agent & generated & tool & \multicolumn{2}{c}{\textbf{PFLOP per dialog}} \\
 & \textbf{params} & & \textbf{turns} & \textbf{tokens} & \textbf{calls} & \textbf{decode} & \textbf{total} \\
\midrule
E2B & 2.3B & 0.483 & 12.4 & 866 & 4.0 & 0.0040 & 0.32 \\
E4B & 4.5B & 0.610 & 12.1 & 852 & 3.9 & 0.0077 & 0.61 \\
E4B {+}GRPO & 4.5B & \textbf{0.697} & 11.7 & 602 & 4.3 & \textbf{0.0054} & \textbf{0.58} \\
26B-A4B & 3.8B & 0.714 & 19.2 & 849 & 13.8 & 0.0065 & 0.86 \\
12B & 12B & 0.690 & 14.4 & 801 & 7.2 & 0.0192 & 2.00 \\
31B & 31B & 0.760 & 12.3 & 589 & 5.5 & 0.0365 & 4.20 \\
MiniMax-M2.7~\cite{minimaxm27} & 10B & 0.804 & 12.8 & 1{,}705 & 6.1 & 0.0341 & 1.86 \\
\bottomrule
\end{tabular}
\caption{Inference cost on the held-out set, 1{,}000 dialogs per model, same
user simulator. Active parameters are the values used in the estimate; PFLOP per dialog $= 2 \times$ active parameters $\times$ tokens, summed over the agent's calls. Decode counts generated tokens only; total also counts every
prompt token, uncached. The decode column is printed to four decimals, since
the trained model's saving over its own base is not visible at three.}
\label{tab:cost}
\end{table}

\subsection{Additional analyses}\label{sec:orderstrict}\label{sec:record}\label{sec:migration}

Three analyses examine how the scoring, the corpus and the training interact:
whether the order of tool calls matters for telling models apart, whether
easier task tiers widen the learnable band, and where the band's tasks go
after training.

\paragraph{Order-strict scoring.} This analysis tests whether the order of
tool calls separates models, or whether scoring the set of calls is already
enough. Both scorings compare the agent's calls against the task's gold chain, the
pre-specified correct calls in their correct order. Deeper chains also mean longer conversations, so the deep slice
doubles as a test of long-context behavior: the model must carry the task
state across more turns to keep the order right. Table~\ref{tab:strict} scores the same held-out ladders twice, once set-based and once under the
in-order gate of Eq.~\ref{eq:seqfrac}, with models, tasks and trajectories held
fixed. The two slices come from the corpus construction of \S\ref{sec:tasks}:
the \emph{shallow slice} holds the first-round families, whose gold chains
have at most 6 actions, and the \emph{deep slice} is the whole corpus, with
gold chains of up to 8 actions. On the shallow slice, ordering does not account for the capability gap. The penalty
stays under 0.04 for
every model while overall scores span 0.48 to 0.80. At that depth the remaining
errors are in arguments and missed steps, so a model calls the right tools in
the right order and still writes the wrong value to the database.

On the deep slice the penalty grows on every model and the in-order match
fraction drops with it, 0.922 to 0.892 for MiniMax-M2.7 and 0.861 to 0.820 for
E4B (Table~\ref{tab:strict}). The penalty grows three to four times on
E2B, E4B, the 12B and 31B references and on MiniMax-M2.7; the 26B-A4B
reference is the exception,
at roughly double. 
Ordering separates models where set-based scoring does not.

The trained model was rescored under the strict gate on all 1{,}000 held-out
tasks at once, so it has no slice rows in Table~\ref{tab:strict}. On that set it
rises from base E4B's 0.590 to 0.679, in step with its set-based gain, and its in-order match fraction
rises from 0.861 to 0.919. The strict gate discounts a correct set of tools that
runs out of order, so a policy that only learned which tools to call could not
raise this score. The gain therefore reflects sequencing skill rather than
leniency in the scorer. It is also where training helped the model with
longer context: the trained model holds the task state across the whole
conversation and keeps the required order.

\begin{table}[htbp]
\centering\small
\begin{tabular}{lcccccc}
\toprule
 & \multicolumn{3}{c}{\textbf{shallow slice (chains $\le 6$)}} & \multicolumn{3}{c}{\textbf{deep slice (chains $\le 8$)}} \\
\cmidrule(lr){2-4}\cmidrule(lr){5-7}
\textbf{model} & \textbf{set} & \textbf{strict} & \textbf{penalty} & \textbf{set} & \textbf{strict} & \textbf{penalty} \\
\midrule
MiniMax-M2.7~\cite{minimaxm27} & 0.804 & 0.769 & $-0.035$ & 0.713 & 0.597 & $-0.116$ \\
31B & 0.760 & 0.728 & $-0.032$ & 0.725 & 0.619 & $-0.106$ \\
26B-A4B & 0.714 & 0.684 & $-0.030$ & 0.587 & 0.531 & $-0.056$ \\
12B & 0.690 & 0.670 & $-0.020$ & 0.613 & 0.523 & $-0.090$ \\
E4B & 0.610 & 0.590 & $-0.020$ & 0.536 & 0.459 & $-0.077$ \\
E2B & 0.483 & 0.462 & $-0.021$ & 0.401 & 0.330 & $-0.071$ \\
\bottomrule
\end{tabular}
\caption{Order-strict rescore of the 1{,}000-task
ladders; penalty is strict minus set-based. The 26B-A4B shallow-slice row was
rescored from all 1{,}000 held-out episodes of that model's release run, at a
mean in-order match fraction of 0.865. Its deep-slice row comes from a separate
1{,}000-task deep-slice run of the same model; 949 of those episodes recorded a
reward and were scored, at a mean in-order match fraction of 0.828.}
\label{tab:strict}
\end{table}

\paragraph{Graded families.} This analysis asks whether adding easier
versions of hard tasks widens the learnable band. The answer is that band
width follows the difficulty mix of the whole corpus. Graded
families step a request from an easy version to a harder one, and they make up
just under 10\% of the corpus. The effect is measured on E2B, the
smallest model, because easier tiers would widen the weakest model's band
first. At that share they leave E2B's learnable share on the deep slice
unchanged: 20.5\% before, 20.4\% after. A larger change in the mix
does move the band. The second generation round shifted the mix toward edge
families, which rise from .266 of the shallow slice to about 40\% of the whole
corpus, and a shift of that size is what sets the band. Widening the band
therefore takes a proportional change in the difficulty mix. That is why the
training sample is drawn with the first-round mix, whose band was measured.

\paragraph{Band migration under training.} This analysis checks where the tasks in the learnable band went after
training: whether they moved into always-pass or stayed in the band. Table~\ref{tab:migration} re-runs the
two-trial evaluation on base and trained E4B together, on the full held-out set
and with the same simulated customer. Over the two trials the base scores 0.595
and the trained model 0.694. Trial 2 on its own returns 0.690, so the result
does not rest on one lucky trial. Always-fail falls about six points and
always-pass rises fourteen. The gap between those two movements is the band,
which narrows from 26.0\% to 17.7\%. Tasks leave the band for always-pass faster
than always-fail refills it.

\begin{table}[htbp]
\centering\small
\begin{tabular}{lccc}
\toprule
\textbf{model} & \textbf{always-fail} & \textbf{learnable} & \textbf{always-pass} \\
\midrule
base E4B & 27.5 & 26.0 & 46.5 \\
{+}GRPO & 21.8 & 17.7 & 60.5 \\
\bottomrule
\end{tabular}
\caption{Trial-outcome split of the 1{,}000-task held-out set, 2 trials, shares
in \%, before and after GRPO; both rows are measured in the same pair of runs.
The base row is the E4B row of Table~\ref{tab:band}, the same
measurement. The trained row is the selected step-180 checkpoint.}
\label{tab:migration}
\end{table}  The ladder predicted the
migration, the falling all-fail share showed it step by step
(Figure~\ref{fig:curves}, third panel), and Table~\ref{tab:migration} shows
where the tasks ended up. Tasks pass through the band instead of staying in it, so during training the
band shrinks faster than new tasks enter it. The band is not empty afterwards: 17.7\% of the held-out corpus still splits across
trials, but a wider band would have to be built into the corpus.

Together, the three analyses give a consistent picture. Ordering separates models only on deep
chains, and easy
tiers help only in proportion to their share of the corpus. E4B's gains landed
on the axes the band analysis said had the most room, and the band that supplied
them is measurably smaller afterwards.

\section{Discussion and Analysis}\label{sec:discussion}\label{sec:complementarity}\label{sec:aba-discussion}

The two studies share the setting of \S\ref{sec:setting}, but they use
different training signals and different evaluation evidence. This discussion
therefore looks first at how preference optimization changes individual model
behaviors, and then at how reinforcement learning affects complete tool-use
trajectories.

The base E4B model already selects tools correctly and constructs valid
\label{sec:disc-dpo}
arguments (\S\ref{sec:aba-pairs}); its failures are behavioral rather than a
missing capability. That makes it a natural fit for post-training that steers
behavior the model already has instead of teaching it from scratch. We chose
DPO over a full reward-model-and-policy loop because the target behavior is
well defined and the failure modes are discrete. A single supervised objective
over preference pairs is enough, and it avoids training and maintaining a
separate reward model (\S\ref{sec:aba-training}). Within DPO, the source of the
preferred response matters. Teacher-sourced preferred responses let the
objective exploit surface shortcuts such as length, discourse markers and
format. The objective then pushes the rejected response down instead of pulling
the preferred one up, and the reward margin grows without bound
(\S\ref{sec:aba-source}). With self-rephrased preferred responses the base model
revises its own output under a rubric. Both sides of the pair then stay inside
the policy's support, which forces the objective to rank them on quality.
This agrees with the finding that on-policy preference data outperforms
higher-quality off-policy data.

The results (\S\ref{sec:aba-results}) support this reasoning, and they
split sharply between what improved and what did not. The behavioral axes (social
engineering, out-of-scope refusal, third-party access) improved sharply, because
the base model already understands these categories but was not applying them
consistently. The capability axes (multitool chains, wrong-info correction)
barely moved, because a preference signal cannot teach reasoning patterns the
model does not yet have. The same split explains why the capability domain rose
from 68\% to 90\% while the banking-task domains moved little. DPO is therefore
the right tool for behavioral alignment in a regulated setting, and closing
capability gaps is likely to need a complementary supervised stage on correct
trajectories.

The improvements are visible axis by axis. Out-of-scope refusal rises from
52\% to 80\%, credentials and inappropriate content reach 100\%, and asking
rather than guessing when information is missing rises from 38\% to 46\%
(Table~\ref{tab:aba-axis}). The one axis that requires tools to be composed
barely moves: multitool chains go from 20\% to 21\%.

Three caveats qualify that reading; none changes the direction of the gain.

\begin{itemize}[leftmargin=*,itemsep=3pt,topsep=4pt]
\item \textbf{Mechanism evidence.} The two constructions differ in the source
of the preferred response, but the reported evaluation does not isolate that
choice. Any difference between teacher-sourced and self-rephrased preference
pairs therefore remains an open question rather than a measured effect.
\item \textbf{Unattributed checkpoint.} The judged tables score a single trained
model without stating which construction of the preferred response produced it,
so the teacher-versus-self comparison is unsettled. The gain belongs to
preference optimization on pairs constructed from failures as a whole, and not
to either construction in particular.
\item \textbf{Single judge, single run.} The reasoning-and-quality tier is
decided by one judge, with no agreement audit of the kind the verifiable reward
gets. The numbers are point estimates from one evaluation, without intervals, so
the finding rests on the size and direction of the achievability gain.
\end{itemize}

The preference-optimization study looks at individual responses within a
conversation. The reinforcement-learning study asks a related question: whether
the model can complete full banking tasks when the outcome depends on which
tools are called, with which arguments, and in what order.

\textbf{Reinforcement learning improved E4B's task execution and reduced its
serving cost:}
\label{sec:disc-grpo}
edge cases rose from 0.509 to 0.718, sequencing from 0.655 to 0.713, and tool
coverage from 0.487 to 0.526 (\S\ref{sec:results-grpo}). All of these
measure which tool is called, with which arguments, and in what order.  A
gradient comes only from tasks whose outcome varies across rollouts, and
training moved those tasks through the learnable band and into always-pass,
rather than leaving them in the band (\S\ref{sec:migration}).

Key points of the reinforcement-learning study:

\begin{itemize}[leftmargin=*,itemsep=3pt,topsep=4pt]
\item \textbf{Past a larger model.} E4B reaches 0.697 held-out reward, above the
base 12B reference's 0.690, at close to a third of the effective
parameters.
\item \textbf{Real sequencing skill.} The gain survives the order-strict gate
(0.590 to 0.679), and the in-order match fraction rises from 0.861 to 0.919.
The model is calling the right tools in the right order, rather than only
choosing a better set of them.
\item \textbf{Cheaper to serve.} 29\% fewer generated tokens per dialog and
0.58 PFLOP per dialog against the 12B rung's 2.00; training made the model
cheaper than its own base on both decode and total compute.
\item \textbf{General capability preserved.} Six public benchmarks move only
slightly, in both directions, so the banking gain left the model's existing
capability intact.
\item \textbf{A verified signal.} The reward that produced the gain agrees with
an independent reference scorer on 97.2\% of reward mass, and the training
dynamics match the learnable-band analysis that predicted them in advance.
\end{itemize}

\section{Conclusion}\label{sec:conclusion}

This paper trained a small open-weight model to act on an Indian retail
banking account, in a way a bank can verify and run on its own hardware. The
work followed a fixed order. We began with five use cases, then built the
scenarios and tools that make them executable, then a simulated bank whose
scoring is checked against an independent reference, and only then trained the
model. Every claim in the paper was measured inside that environment.

Two post-training routes were run on the same 4.5B-parameter model.
Preference pairs constructed from the model's own failures improved its
conduct, with the largest gains on adversarial and out-of-scope requests. A
verifiable reward improved its task execution. The trained model completes tool
sequences at a level above a base model nearly three times its size, and
it does so with fewer generated tokens.

E4B was chosen from the six models measured for two reasons: it is the
smallest model that handles the complexity of the banking tasks in this
corpus, and its footprint is the cheapest to serve inside a bank. The model
below it fails a large share of the tasks outright, while each larger model
adds less than the one before it and costs more to run. The choice therefore
pairs enough capability for the full task set with the lowest serving cost,
and the results confirm it.

This matters directly for Indian retail banking. Several of the five domains are
Indian in their particulars: government scheme eligibility across central and
state programs, insurance claims read against RBI guidelines, TDS on deposit
interest. The behavior the model learns is a requirement of the regulatory
environment, and the tool sequences it learns are the ones those products
need. A model of this size can meet that standard on both counts while
remaining deployable inside bank-controlled infrastructure.

The environment is reusable beyond this model. A task corpus with gold
chains, a replayable environment, and a reward audited against a reference
scorer let every result here be re-run. The same construction, with a different
tool catalog and different use cases, applies to other regulated domains.

\clearpage
\section*{Contributors}

\noindent\textbf{NPCI AI Research Team.}\\
Aman Kumar, Asit Desai, Chandra Bhushan, Harsh Sharma, Harshit
Bhushan, Hrithik Kadam, Keyur Doshi, Kolisetty Sai Kapardheeswar, Krishanu
Adhikary, Nadeem Shaik, Navya Prakash, Nitin Kukreja, Prashant Devadiga,
Shamanth MH, Shantanu Pandey, Suvradip Paul, and Yatharth Dedhia.

\bibliographystyle{unsrt}
\bibliography{refs}

\end{document}